\documentclass[11pt,letterpaper]{article}
\usepackage{emnlp2016}
\usepackage{times}
\usepackage{latexsym}
\usepackage{amsmath,amssymb} 

\usepackage{breakcites}
\usepackage{graphicx}
\usepackage{float}
\usepackage[font=small]{caption}
\usepackage[font=small]{subcaption}
\usepackage{amsthm}
\usepackage{textcomp}
\usepackage{stmaryrd}
\usepackage{upgreek}
\usepackage{bm}
\usepackage{cases}
\usepackage{mathtools}
\usepackage{multirow}
\usepackage{rotating}
\usepackage{booktabs}
\usepackage{slashbox} 
\usepackage{enumerate}
\usepackage{enumitem}
\usepackage[olditem,oldenum]{paralist}
\usepackage{alltt}
\usepackage{listings}
\belowdisplayskip \abovedisplayskip

\newlength{\sectionReduceTop}
\newlength{\sectionReduceBot}
\newlength{\subsectionReduceTop}
\newlength{\subsectionReduceBot}
\newlength{\abstractReduceTop}
\newlength{\abstractReduceBot}
\newlength{\captionReduceTop}
\newlength{\captionReduceBot}
\newlength{\subsubsectionReduceTop}
\newlength{\subsubsectionReduceBot}

\newlength{\eqnReduceTop}
\newlength{\eqnReduceBot}

\newlength{\horSkip}
\newlength{\verSkip}

\newlength{\figureHeight}
\usepackage{url}
\usepackage{xspace}
\usepackage{color}
\usepackage{afterpage}
\usepackage{framed}
\usepackage{fancybox}
\usepackage[normalem]{ulem}
\usepackage{dsfont}

\newcommand{\wordtwovec}{word2vec\xspace}

\newcommand{\ie}{\emph{i.e.}\xspace}
\newcommand{\eg}{\emph{e.g.}\xspace}
\newcommand{\figref}[1]{Figure~\ref{#1}}
\newcommand{\secref}[1]{Section~\ref{#1}}
\newcommand{\mapmap}{\textsc{Indep}\xspace}
\newcommand{\cascade}{\textsc{Cascade}\xspace}
\newcommand{\domainadapt}{\textsc{Domain Adaptation}\xspace}

\newcommand{\divmbest}{DivMBest\xspace}

\newcommand{\mediator}{\textsc{Mediator}\xspace}

\newcommand{\xb}{\mathbf{x}}

\newcommand{\yb}{\mathbf{y}}
\newcommand{\zb}{\mathbf{z}}

\newcommand{\ygt}{\yb^{gt}}
\newcommand{\zgt}{\zb^{gt}}

\newcommand{\yhat}{\hat{\yb}}

\newcommand{\Yb}{\mathbf{Y}}
\newcommand{\Zb}{\mathbf{Z}}

\newcommand{\loss}{\ell}
\newcommand{\rloss}{\mathcal{L}}

\newcommand{\wb}{\mathbf{w}} 

\newcommand{\feat}{\bfgreek{phi}}

\newcommand{\oracle}{\texttt{oracle}\xspace}

\newcommand{\train}{$\mathtt{train}$\xspace}
\newcommand{\val}{$\mathtt{val}$\xspace}
\newcommand{\test}{$\mathtt{test}$\xspace}

\newcommand{\beq}{\begin{eqnarray*}}
\newcommand{\eeq}{\end{eqnarray*}}
\newcommand{\beqn}{\begin{eqnarray}}
\newcommand{\eeqn}{\end{eqnarray}}
\newcommand{\ben}{\begin{enumerate}}
\newcommand{\een}{\end{enumerate}}
\newcommand{\bit}{\begin{itemize}}
\newcommand{\eit}{\end{itemize}}
\newcommand{\hide}[1]{}

\newcommand{\argmax}{\mathop{\mathrm{argmax}}}

\newcommand{\calM}{\mathcal{M}}

\newcommand{\calX}{\mathcal{X}}
\newcommand{\calY}{\mathcal{Y}}
\newcommand{\calZ}{\mathcal{Z}}

\newcommand{\tableScale}{0.9}

\makeatletter
\DeclareRobustCommand\onedot{\futurelet\@let@token\@onedot}
\def\@onedot{\ifx\@let@token.\else.\null\fi\xspace}

\def\eg{\emph{e.g}\onedot} 
\def\ie{\emph{i.e}\onedot} 
\def\st{s.t\onedot}

\def\wrt{w.r.t\onedot} 

\makeatother

\newcommand{\eqnref}[1]{(\ref{#1})}

\makeatletter
\newcommand{\bfgreek}[1]{\bm{\@nameuse{#1}}}
\makeatother

     {\begin{list}{}%
             {\setlength{\leftmargin}{#1}}%
             \item[]%
     }
{\end{list}}

\emnlpfinaltrue

\newcommand\blfootnote[1]{
  \begingroup
  \renewcommand\thefootnote{}\footnote{#1}
  \addtocounter{footnote}{-1}
  \endgroup
}

\begin{document}

\title{Resolving Language and Vision Ambiguities Together: Joint \newline Segmentation \& Prepositional Attachment Resolution in Captioned Scenes} 

\author{Gordon Christie$^{1,*}$, Ankit Laddha$^{2,*}$, Aishwarya Agrawal$^1$, Stanislaw Antol$^1$ \\ {\bf Yash Goyal}$^1$, {\bf Kevin Kochersberger}$^1$, {\bf Dhruv Batra}$^{3,1}$ \\ $^1$Virginia Tech \quad $^2$Carnegie Mellon University \quad $^3$Georgia Institute of Technology \\ {\tt ankit1991laddha@gmail.com} \\ {\tt \{gordonac,aish,santol,ygoyal,kbk,dbatra\}@vt.edu}}

\date{}

\def\emnlppaperid{959}

\maketitle


\vspace{\abstractReduceTop}
\begin{abstract} 
\vspace{\abstractReduceBot}
We present an approach to simultaneously perform semantic segmentation and prepositional phrase attachment resolution for captioned images. Some ambiguities in language cannot be resolved without simultaneously reasoning about an associated image. If we consider the sentence ``I shot an elephant in my pajamas'', looking at language alone (and not using common sense), it is unclear if it is the person or the elephant wearing the pajamas or both. Our approach produces a diverse set of plausible hypotheses for both semantic segmentation and prepositional phrase attachment resolution that are then jointly reranked to select the most consistent pair. We show that our semantic segmentation and prepositional phrase attachment resolution modules have complementary strengths, and that joint reasoning produces more accurate results than any module operating in isolation. Multiple hypotheses are also shown to be crucial to improved multiple-module reasoning. Our vision and language approach significantly outperforms the Stanford Parser~\cite{de2006generating} by 17.91\% (28.69\% relative) and 12.83\% (25.28\% relative) in two different experiments. We also make small improvements over DeepLab-CRF \cite{chen14semantic}. 
\end{abstract} 
\vspace{\abstractReduceBot}


\vspace{\sectionReduceTop}
\section{Introduction}
\vspace{\sectionReduceBot}
\label{sec:intro}

\blfootnote{* Denotes equal contribution}

\begin{figure}[ht!]
  \centering
  \includegraphics[width=0.96\columnwidth]{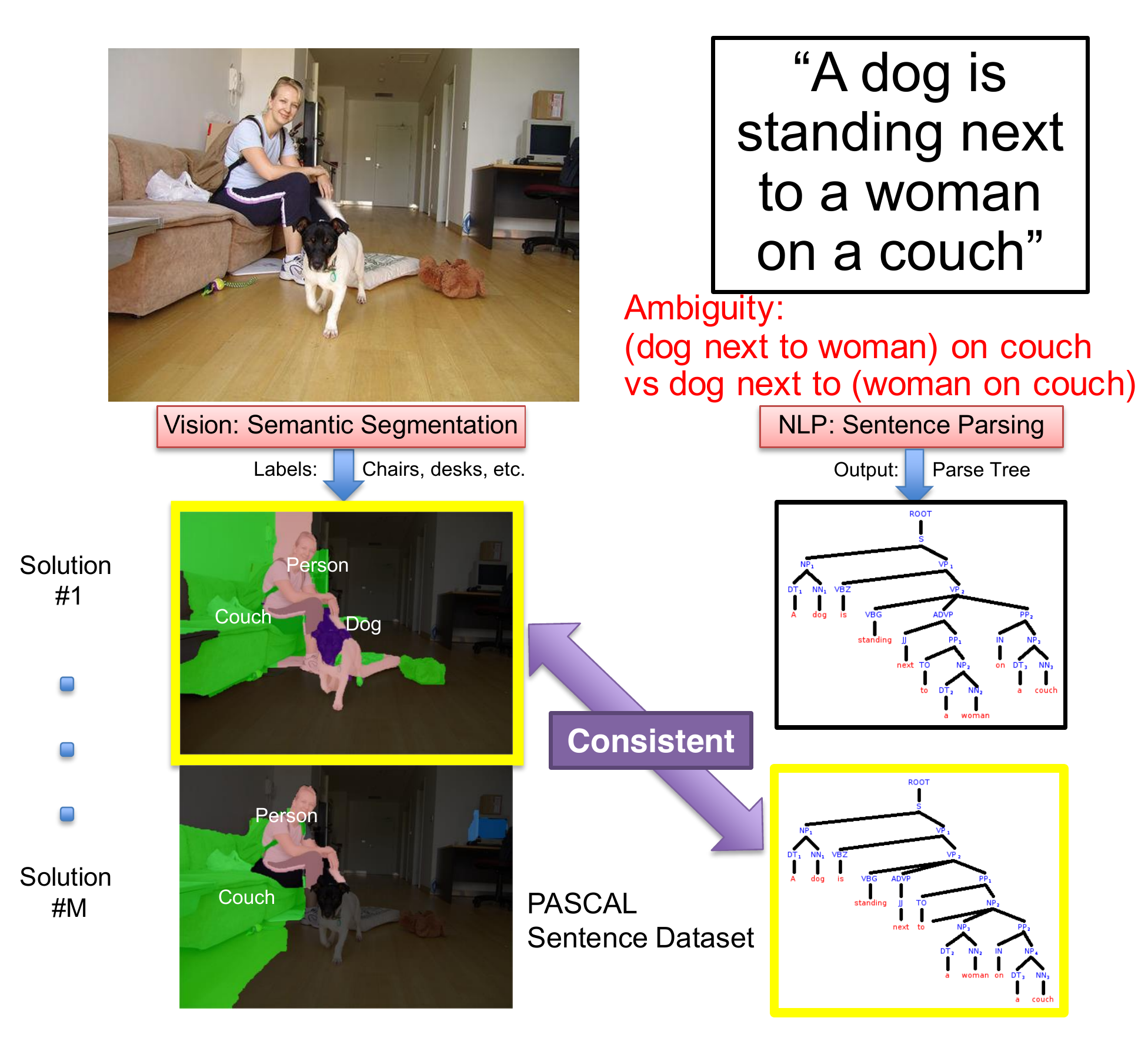}
   \caption{{\small Overview of our approach. We propose a model for simultaneous 2D semantic segmentation and prepositional phrase attachment resolution by reasoning about sentence parses. The language and vision modules each produce $M$ diverse hypotheses, and the goal is to select a pair of consistent hypotheses.  In this example the ambiguity to be resolved from the image caption is whether the dog is standing on or next to the couch.  Both modules benefit by selecting a pair of compatible hypotheses.}}
  \vspace{-10pt}
    \label{fig:overview}
  \vspace{\captionReduceBot}
\end{figure}

Perception and intelligence problems are hard. Whether we are interested in understanding an image or a sentence, 
our algorithms must operate under tremendous levels of ambiguity. 
When a human reads the sentence ``I eat sushi with tuna'', it is clear that the preposition phrase ``with tuna'' modifies ``sushi'' and not the act of eating, but this may be ambiguous to a machine.    
This problem of determining whether a prepositional phrase (``with tuna'') modifies a noun phrase (``sushi") or verb phrase (``eating") is formally known as Prepositional Phrase Attachment Resolution (PPAR)~\cite{ratnaparkhi1994maximum}. 
Consider the captioned scene shown in \figref{fig:overview}. The caption ``A dog is standing next to a woman on a couch'' exhibits a PP attachment ambiguity -- ``(dog next to woman) on couch'' vs ``dog next to (woman on couch)''.
It is clear that having access to image segmentations can help resolve this ambiguity, and having access to the correct PP attachment can help image segmentation.

There are two main roadblocks that keep us from writing a single unified model (say a graphical model) to perform both tasks: 
(1) Inaccurate Models --  empirical studies~\cite{meltzer_iccv05,szeliski_pami08,kappes_cvpr13} have repeatedly found 
that models are often inaccurate and miscalibrated -- 
their ``most-likely'' beliefs are placed on solutions far from the ground-truth. 
(2) Search Space Explosion -- jointly reasoning
about multiple modalities is difficult due to the combinatorial explosion of search space 
(\{exponentially-many segmentations\} $\times$ \{exponentially-many sentence-parses\}).

\textbf{Proposed Approach and Contributions.}
In this paper, we address the problem of simultaneous object segmentation (also called semantic segmentation) and PPAR in captioned scenes. To the best of our knowledge this is the first paper to do so.

Our main thesis is that a set of diverse plausible hypotheses 
can serve as a concise interpretable summary of uncertainty in vision and language `modules' 
(What does the semantic segmentation module see in the world? What does the PPAR module describe?)
and form the basis for tractable joint reasoning 
(How do we reconcile what the semantic segmentation module sees in the world with how the PPAR module describes it?).

Given our two modules with $M$ hypotheses each, how can we 
integrate beliefs across the segmentation and sentence parse modules to pick the best 
pair of hypotheses?
Our key focus is \emph{consistency} -- correct hypotheses from different modules 
will be correct in a consistent way, but incorrect hypotheses will be incorrect in incompatible ways.
Specifically, we develop a \mediator model that scores pairs for consistency 
and searches over all $M^2$ pairs to pick the highest scoring one. 
We demonstrate our approach on three datasets -- ABSTRACT-50S \cite{rama50S}, PASCAL-50S, and PASCAL-Context-50S~\cite{mottaghi_cvpr14}.
We show that our vision+language approach significantly outperforms the Stanford Parser~\cite{de2006generating} by 20.66\% (36.42\% relative) for ABSTRACT-50S, 17.91\% (28.69\% relative) for PASCAL-50S, and by 12.83\% (25.28\% relative) for PASCAL-Context-50S. We also make small but consistent improvements over DeepLab-CRF~\cite{chen14semantic}.  

\vspace{\sectionReduceTop}
\section{Related Work}
\vspace{\sectionReduceBot}
\label{sec:related}

Most works at the intersection of vision and NLP tend to be `pipeline' systems, where vision tasks take 1-best inputs from NLP (\eg, sentence parsings) without trying to improve NLP performance and vice-versa.
For instance, \newcite{fidler_cvpr13} use prepositions to improve object segmentation and scene classification, but only consider
the most-likely parse of the sentence and do not  resolve ambiguities in text. 
Analogously, \newcite{yatskar2014see} investigate the role of object, attribute, and action classification annotations for generating human-like descriptions. 
While they achieve impressive results at generating descriptions, they assume perfect vision modules to generate sentences.
Our work uses current (still imperfect) vision and NLP modules to reason about images and provided captions, and simultaneously improve both vision and language modules.
Similar to our philosophy, an earlier work by \newcite{barnard2005word} used images to help disambiguate word senses (\eg piggy banks vs snow banks).
In a more recent work, \newcite{gella2016unsupervised} studied the problem of reasoning about an image and a verb, where they attempt to pick the correct sense of the verb that describes the action depicted in the image.
\newcite{berzak2016you} resolve linguistic ambiguities in sentences coupled with videos that represent different interpretations of the sentences.
Perhaps the work closest to us is \newcite{kong_cvpr14}, who leverage information from 
an RGBD image and its sentential description 
to improve 3D semantic parsing and 
resolve ambiguities related to co-reference resolution in the sentences (\eg, what ``it'' refers to). 
We focus on a different kind of ambiguity -- the Prepositional Phrase (PP) attachment resolution. 
In the classification of parsing ambiguities, co-reference resolution is considered a 
discourse ambiguity \cite{coref_res} (arising out of two different words across sentences for the same object), 
while PP attachment is considered a syntactic ambiguity (arising out of multiple valid sentence structures) 
and is typically considered much more difficult to resolve \cite{Bach_ambiguity,Davis_ambiguity}. 

A number of recent works have studied problems at the intersection of vision and language, such as Visual Question Answering~\cite{antol2015vqa,geman,Malinowski_2015_ICCV}, Visual Madlibs~\cite{yu2015visual}, and image captioning~\cite{captioning_google,captioning_msr}. Our work falls in this domain with a key difference that we 
produce \emph{both} vision and NLP outputs. 

Our work also has similarities with works on `spatial relation learning'~\cite{malinowski2014pooling,lan2012image}, \ie learning a visual representation for noun-preposition-noun triplets (``car on road''). While our approach can certainly utilize such spatial relation classifiers if available, the focus of our work is different. Our goal is to improve semantic segmentation and PPAR by jointly reranking segmentation-parsing solution pairs. Our approach implicitly learns spatial relationships for prepositions (``on'', ``above'') but these are simply emergent latent representations that help our reranker pick out the most consistent pair of solutions. 

Our work utilizes a line of work \cite{batra_eccv12, batra_uai12, prasad_nips14} on producing diverse plausible solutions from probabilistic models, which has been successfully applied to a number of problem domains \cite{rivera_aistats13, yadollahpour_cvpr13, gimpel_emnlp13, premachandran_cvpr14, sun_cvpr15, ahmed_iccv15}.

\vspace{\sectionReduceTop}
\section{Approach}
\vspace{\sectionReduceBot}
\label{sec:model}

In order to emphasize the generality of our approach, and to show that our approach is compatible with a wide class of implementations of semantic segmentation and PPAR modules%
, we present our approach 
with the modules abstracted as ``black boxes'' that satisfy 
a few general requirements and minimal assumptions.
In \secref{sec:exp}, we describe each of the modules in detail, 
making concrete their respective features, and other details. 

\vspace{\subsectionReduceTop}
\subsection{What is a Module?} 
\vspace{\subsectionReduceBot}

The goal of a module is to take input variables $\xb \in \calX$ 
(images or sentences), and predict output variables 
$\yb \in \calY$ (semantic segmentation) and $\zb \in \calZ$
(prepositional attachment expressed 
in sentence parse).
The two requirements on a module are that it needs to be able to 
produce \emph{scores} $S(\yb | \xb)$ for potential solutions and 
a list of \emph{plausible hypotheses} $\Yb = \{\yb^1, \yb^2, \ldots, \yb^M\}$.

\textbf{Multiple Hypotheses.}
In order to be useful, the set $\Yb$ of hypotheses must provide an accurate 
summary of the score landscape. Thus, the hypotheses should be plausible (\ie, high-scoring) 
and mutually non-redundant (\ie, diverse). 
Our approach (described next) is applicable to any choice of diverse 
hypothesis generators. 
In our experiments, we use the k-best algorithm of \newcite{huang_wpt05} for the sentence parsing module 
and the \divmbest algorithm~\cite{batra_eccv12} for the semantic segmentation module.
Once we instantiate the modules in \secref{sec:exp}, we describe the diverse solution generation in more detail.

\vspace{\subsectionReduceTop}
\subsection{Joint Reasoning Across Multiple Modules}
\label{sec:integrated}
\vspace{\subsectionReduceBot}

We now show how to intergrate information from both 
segmentation and PPAR modules. 
Recall that our key focus is \emph{consistency} -- correct hypotheses from different modules 
will be correct in a consistent 
way, but incorrect hypotheses will be incorrect in incompatible ways. 
Thus, our goal is to search for a pair (semantic segmentation, sentence parsing)
that is mutually consistent.

Let $\Yb = \{\yb^1,\ldots, \yb^M\}$ denote the $M$ semantic segmentation hypotheses  
and $\Zb = \{\zb^1,\ldots, \zb^M\}$ denote the $M$ PPAR hypotheses. 

\begin{figure*}[t!]
\centering
\includegraphics[width=\textwidth]{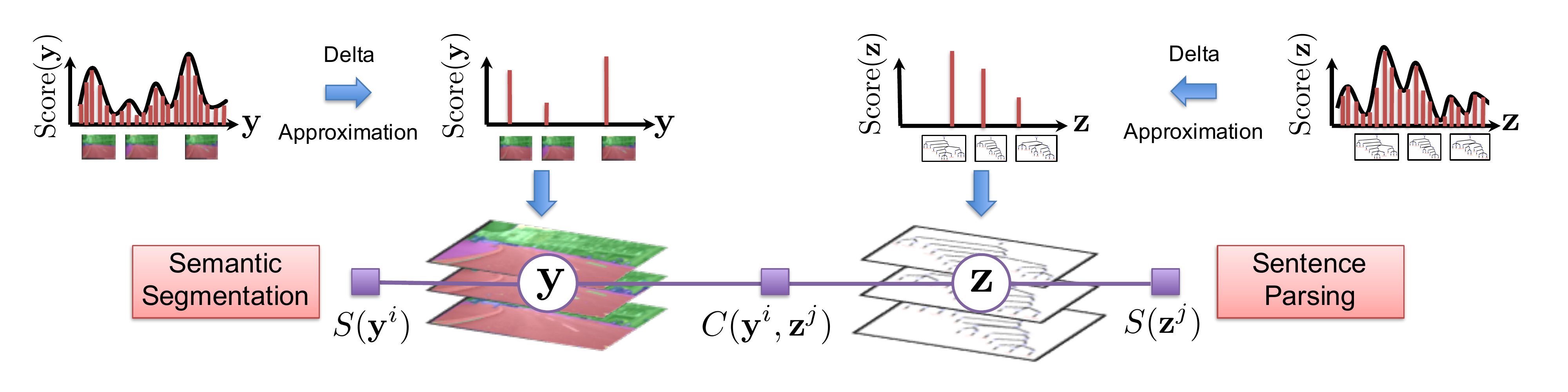}
\caption{{\small Illustrative inter-module factor graph. Each node takes exponentially-many  
or infinitely-many states and we use a `delta approximation' to limit support.} 
}
\label{fig:module_factors}
\vspace{-0.5cm}
\end{figure*}

\textbf{\textsc{mediator} Model.}
We develop a ``mediator'' model that identifies high-scoring hypotheses across modules 
in agreement with each other. Concretely, we can express the \mediator model 
as a factor graph where each node corresponds to a module (semantic segmentation and PPAR).
Working with such a factor graph is typically completely intractable because each node
 $\yb, \zb$
has exponentially-many states (image segmentations, sentence parsing). As illustrated in \figref{fig:module_factors}, in this factor-graph view, the hypothesis sets 
$\Yb, \Zb$ 
can be considered `delta-approximations' for reducing the size of the 
output spaces. 

Unary factors $S(\cdot)$ capture the score/likelihood of each hypothesis provided by the 
corresponding module for the image/sentence at hand. 
Pairwise factors 
$C(\cdot, \cdot)$
represent consistency factors. 
Importantly, since we have restricted each module variables to just $M$ states, 
we are free to capture \emph{arbitrary domain-specific high-order relationships} for consistency, 
without any optimization concerns. 
In fact, as we describe in our experiments, these consistency factors may be 
designed to exploit domain knowledge in fairly sophisticated ways.

\textbf{Consistency Inference.}
We perform exhaustive inference over 
all possible tuples. 
{\small
\begin{align}
\argmax_{i,j \in \{1,\ldots,M\}} \Big\{\calM(\yb^i, \zb^j) = 
S(\yb^i) + S(\zb^j) + C(\yb^i,\zb^j) \Big\}.
\end{align}
}
Notice that the search space with $M$ 
hypotheses each is $M^2$.
In our experiments, we allow each module to take a different value for $M$, 
and typically use around 10 solutions for each module, leading to a mere 100 
pairs, which is easily enumerable. 
We found that even with such a small set, 
at least one of the solutions in the set tends to be \emph{highly accurate}, meaning that the hypothesis 
sets have relatively high recall. 
This shows the power of using a small set of diverse hypotheses. 
For a large $M$, we can exploit a number of standard ideas from the graphical models literature (\eg dual decomposition or belief propagation). In fact, this is one reason we show the factor in \figref{fig:module_factors}; there is a natural decomposition of the problem into modules.

\textbf{Training \textsc{mediator}.}
We can express the \mediator score as 
$ \calM(\yb^i, \zb^j) = \wb^\intercal \feat(\xb,\yb^i, \zb^j)$, 
as a linear function of \emph{score and consistency features}
$\feat(\xb,\yb^i, \zb^j) = [ \phi_{S}(\yb^i); \phi_{S}(\zb^j); \phi_{C}(\yb^i,\zb^j) ]$
, where $\phi_{S}(\cdot)$ are the single-module (semantic segmentation and PPAR module) score features, and $\phi_{C}(\cdot,\cdot)$ are the inter-module consistency features.
We describe these features in detail in the experiments. 
We learn these consistency weights $\wb$ from a dataset annotated with ground-truth 
for the two modules
$\yb, \zb$. 
Let 
$\{\yb^*, \zb^*\}$
denote the \oracle pair, composed of the most accurate 
solutions in the hypothesis sets. 
We learn the \mediator parameters in a discriminative learning fashion
by solving the following Structured SVM problem:
{\small
\begin{subequations}
\label{eq:qp}
\begin{align}
\hspace{-5pt}
\min_{\wb, \xi_{ij}} \quad & \frac{1}{2}\wb^\intercal\wb + C \sum_{ij} \xi_{ij} \label{eq:rerankobjective} \\
\begin{split}
\st \quad & \underbrace{\wb^\intercal \feat(\xb, \yb^*, \zb^*)}_{\text{Score of \oracle tuple}} 
- \underbrace{\wb^\intercal \feat(\xb, \yb^i, \zb^j)}_{\text{Score of any other tuple}} \\
&\ge \underbrace{1}_{\text{Margin}} - \underbrace{\frac{\xi_{ij}}{\rloss(\yb^i, \zb^j)}}_{\text{Slack scaled by loss}}  
\quad\,\, \forall i,j \in \{1,\ldots,M\}.
\label{eqn:rmargin}
\end{split}
\end{align}
\end{subequations}
}
Intuitively, we can see that the constraint \eqnref{eqn:rmargin} tries to maximize the (soft) margin between the
score of the \oracle pair and all other pairs in the hypothesis sets.
Importantly, the slack (or violation in the margin) is scaled by the loss of the tuple. 
Thus, if there are other good pairs not too much worse than the \oracle, 
the margin for such tuples will not be tightly enforced. 
On the other hand, the margin between the \oracle and bad tuples will be very strictly enforced.

This learning procedure requires us to define the loss function
$\rloss(\yb^i, \zb^j)$
, \ie, the cost of predicting a tuple (semantic segmentation, sentence parsing).
We use a weighted average of individual losses:
\begin{align}
\label{eq:weighted_loss}
\rloss(\yb^i, \zb^j) = \alpha \loss(\ygt,\yb^i) + (1-\alpha) \loss(\zgt,\zb^j)
\end{align} 
The standard measure for evaluating semantic segmentation is average Jaccard Index (or Intersection-over-Union)~\cite{pascal-voc},
while for evaluating sentence parses \wrt their prepositional phrase attachment, 
we use the fraction of prepositions 
correctly attached. 
In our experiments, we report results with such a convex combination of module loss functions (for different values of $\alpha$).

\vspace{\sectionReduceTop}
\section{Experiments}
\vspace{\sectionReduceBot}
\label{sec:exp}
We now describe the setup of our experiments, 
provide implementation details of the modules, 
and describe the consistency features. 

\textbf{Datasets.}
Access to rich annotated image + caption datasets is crucial for performing quantitative evaluations. Since this is the first paper to study the problem of joint segmentation and PPAR, no standard datasets for this task exist so we had to curate our own annotations for PPAR on three image caption datasets -- ABSTRACT-50S~\cite{rama50S},  
PASCAL-50S~\cite{rama50S} 
(expands the UIUC PASCAL sentence dataset~\cite{rashtchian2010collecting} from 5 
captions per image to 50), and PASCAL-Context-50S~\cite{mottaghi_cvpr14} (which uses the PASCAL Context image annotations and the same sentences as PASCAL-50S). 
Our annotations are publicly available on the authors' webpages.
To curate the PASCAL-Context-50S PPAR annotations, we first select all sentences that have preposition phrase attachment ambiguities. We then plotted the distribution of prepositions in these sentences. The top 7 prepositions are used, as there is a large drop in the frequencies beyond these. The 7 prepositions are: ``on'', ``with'', ``next to'', ``in front of'', ``by'', ``near'', and ``down''. We then further sampled sentences to ensure uniform distribution across prepositions. We perform a similar filtering for PASCAL-50S and ABSTRACT-50S (using the top-6 prepositions for ABSTRACT-50S). Details are in the appendix. 
We consider a preposition ambiguous if there are at least two parsings where one of the two objects in the preposition dependency 
is the same across the two parsings while the other object is different (\eg (dog on couch) and (woman on couch)).
To summarize the statistics of all three datasets:

\begin{compactenum}
\item \textbf{ABSTRACT-50S}~\cite{rama50S}: 25,000 sentences (50 per image) with 500 images from abstract scenes made from clipart. Filtering for captions containing the top-6 prepositions resulted in 399 sentences describing 201 unique images. These 6 prepositions are: ``with'', `next
to'', ``on top of'', ``in front of'', ``behind'', and ``under''. Overall, there are 502 total prepositions, 406 ambiguous prepositions, 80.88\% ambiguity rate and 60 sentences with multiple ambiguous prepositions.
\item \textbf{PASCAL-50S}~\cite{rama50S}: 50,000 sentences (50 per image) for the images in the UIUC PASCAL sentence dataset~\cite{rashtchian2010collecting}. Filtering for the top-7 prepositions resulted in a total of 30 unique images, and 100 image-caption pairs, where ground-truth PPAR were carefully annotated by two vision + NLP graduate students. Overall, there are 213 total prepositions, 147 ambiguous prepositions, 69.01\% ambiguity rate and 35 sentences with multiple ambiguous prepositions.
\item \textbf{PASCAL-Context-50S}~\cite{mottaghi_cvpr14}: We use images and captions from PASCAL-50S, but with PASCAL Context segmentation annotations (60 categories instead of 21). This makes the vision task more challenging. Filtering this dataset for the top-7 prepositions resulted in a total of 966 unique images and 1,822 image-caption pairs. Ground truth annotations for the PPAR were collected using Amazon Mechanical Turk.  Workers were shown an image 
and a prepositional attachment (extracted from the corresponding parsing 
of the caption) as a phrase (``woman on couch''), and asked if it was correct. A screenshot of our interface 
is available in the appendix. 
Overall, there are 2,540 total prepositions, 2,147 ambiguous prepositions, 84.53\% ambiguity rate and 283 sentences with multiple ambiguous prepositions.
\end{compactenum}

\textbf{Setup.} \textit{Single Module:} We first show that visual features help PPAR by using the ABSTRACT-50S dataset, which contains 
clipart scenes where the extent and position of all the objects in the scene is known. 
This allows us to consider a scenario with a perfect vision system. 

\textit{Multiple Modules:} In this experiment we use imperfect language and vision modules, and show improvements on the 
PASCAL-50S and 
PASCAL-Context-50S datasets. 

\textbf{Module 1: Semantic Segmentation (SS) $\yb$.} 
We use DeepLab-CRF~\cite{chen14semantic} and \divmbest~\cite{batra_eccv12}
to produce $M$ diverse segmentations of the images. To evaluate we use 
image-level class-averaged Jaccard Index.

\textbf{Module 2: PP Attachment Resolution (PPAR) $\zb$.}
We use a recent version (v3.3.1; released 2014) of the PCFG Stanford parser module~\cite{de2006generating, huang_wpt05} to produce $M$ 
parsings of the sentence.
In addition to the parse trees, the module can also output \emph{dependencies}, 
which make syntactical relationships more explicit. 
Dependencies come in the form \emph{dependency\_type($word_1$, $word_2$)}, 
such as the preposition dependency \emph{prep\_on(woman-8, couch-11)} 
(the number indicates the word position in sentence). 
To evaluate, 
we count the percentage of preposition attachments that the parse gets correct. 

\textbf{Baselines:} 

\begin{itemize}[noitemsep,nolistsep]
\item \mapmap. In our experiments, we compare our proposed approach (\mediator) 
to the highest scoring solution predicted independently from each module. For semantic segmentation this is the output of DeepLab-CRF~\cite{chen14semantic} and for the PPAR module this is the 1-best output of the Stanford Parser~\cite{de2006generating, huang_wpt05}.
Since our hypothesis lists are generated by greedy M-Best algorithms, this corresponds to predicting the $(\yb^1, \zb^1)$ tuple. This comparison establishes the importance of joint reasoning.
To the best of our knowledge, there is no existing (or even natural) joint model to compare to.
\item \domainadapt.
We learn a reranker on the parses. Note that domain adaptation is only needed for PPAR since the Stanford parser is trained on Penn Treebank (Wall Street Journal text) and not on text about images (such as image captions). Such domain adaptation is not necessary for semantic segmentation.
This is a competitive single-module baseline. Specifically, we use the \emph{same} parse-based features as our approach, and learn a reranker over the the $M_z$ parse trees ($M_z = 10$). 
\end{itemize}

Our approach (\mediator) significantly outperforms both baselines. The improvements over \mapmap show that joint reasoning produces
more accurate results than any module (vision or language) operating in isolation. 
The improvements over \domainadapt establish the source of improvements is indeed vision, and not the reranking step. Simply adapting the parse from its original training domain (Wall Street Journal) to our domain (image captions) is not enough.

\textbf{Ablative Study.} Ours-\cascade: 
This ablation studies the importance of multiple hypothesis. 
For each module (say $\yb$), 
we feed the single-best output of the other module $\zb^1$ as input. Each module learns 
its own weight $\wb$ using \emph{exactly the same} consistency features 
and learning algorithm as \mediator and predicts one of the plausible hypotheses 
$\yhat^{\cascade} = \argmax_{\yb \in \Yb} \wb^\intercal \feat(\xb,\yb,\zb^1)$.
This ablation of our system is similar to \cite{heitz_nips08} and helps us in 
disentangling the benefits of multiple hypothesis and joint reasoning. 

Finally, we note that Ours-\cascade can be viewed as special cases of \mediator.
Let \mediator-$(M_\yb, M_\zb)$ denote our approach run with $M_\yb$ hypotheses 
for the first module and $M_\zb$ for the second. Then \mapmap corresponds to \mediator-$(1,1)$ and \cascade corresponds to 
predicting the $\yb$ solution from
\mediator-$(M_\yb,1)$ 
and the $\zb$ solution from
\mediator-$(1,M_\zb)$. To get an upper-bound 
on our approach, 
we report \oracle, the accuracy of the most accurate tuple in $10 \times 10$ tuples. 

Our results are presented where \mediator was trained with equally weighted loss ($\alpha = 0.5$), but we provide additional results for varying values of $\alpha$ in the appendix.

\textbf{\mediator and Consistency Features.}
Recall that we have two types of features -- (1) score features $\phi_{S}(y^i)$ and $\phi_{S}(z^j)$, which try to capture how likely solutions $y^i$ and $z^j$ are respectively, and (2) consistency features $\phi_{C}(y^i, z^j)$, which capture how consistent the PP attachments in $z^j$ are with the segmentation in $y^i$. For each ($object_1$, \emph{preposition}, $object_2$) in $z^j$, we compute 6 features between $object_1$ and $object_2$ segmentations in $y^i$. 
Since the humans writing
the captions may use multiple synonymous words (\eg dog, puppy) for
the same visual entity, we use
\wordtwovec~\cite{mikolov2013efficient} similarities to map the nouns in the sentences to the corresponding
dataset categories.

\begin{compactitem}

\item \textbf{Semantic Segmentation Score Features ($\phi_{S}(y^i)$) (2-dim)}: We use ranks 
and solution scores from DeepLab-CRF~\cite{chen14semantic}.

\item \textbf{PPAR Score Features ($\phi_{S}(z^i)$) (9-dim)}: We use ranks
and the log probability of parses from~\cite{de2006generating}, 
and 7 binary indicators for PASCAL (6 for ABSTRACT-50S) denoting which prepositions are present in the parse.

\item \textbf{Inter-Module Consistency Features (56-dim)}: 
For each of the 7 prepositions, 8 features are calculated:
\begin{compactitem}
\item One feature is the Euclidean distance between the center of the segmentation masks of the two objects connected by the preposition. These two objects in the segmentation correspond to the categories with which the soft similarity of the two objects in the sentence is highest among all PASCAL categories. 
\item Four features capture $\max$\{0, (normalized -directional-distance)\}, where directional-distance measures above/below/left/right displacements between the two objects in the segmentation, and normalization involves dividing by height/width.
\item One feature is the ratio of sizes between object$_1$ and object$_2$ in the segmentation.
\item Two features capture the \wordtwovec similarity between the two objects in PPAR (say `puppy' and `kitty') with their most similar PASCAL category (say `dog' and `cat'), where these features are 0 if the categories are not present in segmentation.
\end{compactitem}

A visual illustration for some of these features for PASCAL can be seen in \figref{fig:visual_feats_main}.
\begin{figure}[!t]
	\centering
	\includegraphics[width=0.76\columnwidth]{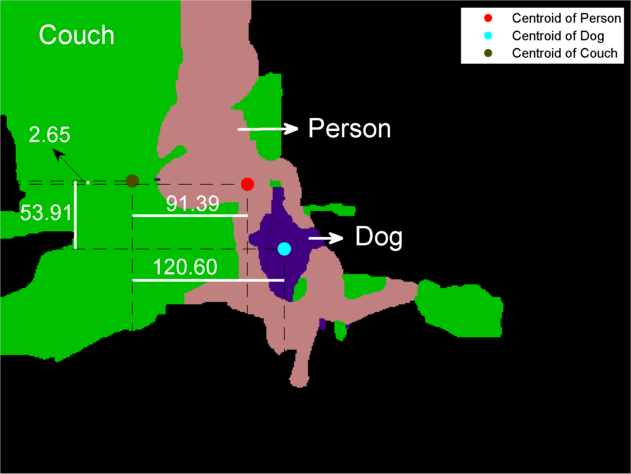}
	\caption{
	{ \small
	Example on PASCAL-50S (``A dog is standing next to a woman on a couch.'').
	The ambiguity in this sentence ``(dog next to woman) on couch'' vs ``dog next to (woman on couch)''. We calculate the horizontal and vertical distances between the segmentation centers of ``person'' and ``couch'' and between the segmentation centers of ``dog'' and ``couch''.  We see that the ``dog'' is much further below the couch ($53.91$) than the woman ($2.65$). So, if the \mediator model learned that ``on'' means the first object is above the second object, we would expect it to choose the ``person on couch'' preposition parsing. \\
	}}
	\label{fig:visual_feats_main}
    \vspace{-0.9cm}
\end{figure}
In the case where an object parsed from $z^j$ is not present in the segmentation $y^i$, the distance features are set to 0. 
The ratio of areas features (area of smaller object / area of larger object) are also set to 0 assuming that the smaller object is missing. In the case where an object has two or more connected components in the segmentation, the distances are computed w.r.t. the centroid of the segmentation and the area is computed as the number of pixels in the union of the instance segmentation masks. 
We also calculate 20 features for PASCAL-50S and 59 features for PASCAL-Context-50S that capture that consistency between $y^i$ and $z^j$, in terms of presence/absence of PASCAL categories. 
For each noun in PPAR we compute its 
\wordtwovec 
similarity with all PASCAL categories. For each of the PASCAL categories, the feature is the sum of similarities (with the PASCAL category) over all nouns if the category is present in segmentation, and is -1 times the sum of similarities over all nouns otherwise.
This feature set was not used for ABSTRACT-50S, since these features were intended to help improve the accuracy of the semantic segmentation module. For ABSTRACT-50S, we only use the 5 distance features, resulting in a 30-dim feature vector.

\end{compactitem}

\vspace{\subsectionReduceTop}
\subsection{Single-Module Results}
\vspace{\subsectionReduceBot}
We performed a 10-fold cross-validation on the ABSTRACT-50S dataset to pick $M$ (=10) and the weight on the hinge-loss for \mediator ($C$). 
The results are presented in Table~\ref{table:sspp_single}.
Our approach significantly outpeforms 1-best outputs of the Stanford Parser \cite{de2006generating} by 20.66\% (36.42\% relative). This shows
 a need for diverse hypotheses and reasoning about visual features when picking a sentence parse. \oracle denotes the best achievable performance using these 10 hypotheses.

\begin{table}[!h]
\centering
\setlength{\tabcolsep}{8pt}
{\small
\scalebox{\tableScale}{
\begin{tabular}{ccccc}
\toprule  
Module  & \begin{tabular}[x]{@{}c@{}}Stanford \\ Parser \end{tabular} 
& \begin{tabular}[x]{@{}c@{}}Domain \\ Adaptation \end{tabular} &  Ours & \oracle \\
\midrule
PPAR   & 56.73 & 57.23 & \textbf{77.39} & 97.53 \\
\bottomrule
\end{tabular}
}
}
\caption{{\small Results on our subset of ABSTRACT-50S.}}
\label{table:sspp_single}
\vspace{-0.3cm}
\end{table}

\subsection{Multiple-Module Results}

\begin{figure*}[ht!]
    \centering
   \begin{subfigure}[b]{0.24\textwidth}
   		  \centering
		  \includegraphics[height=4.3cm]{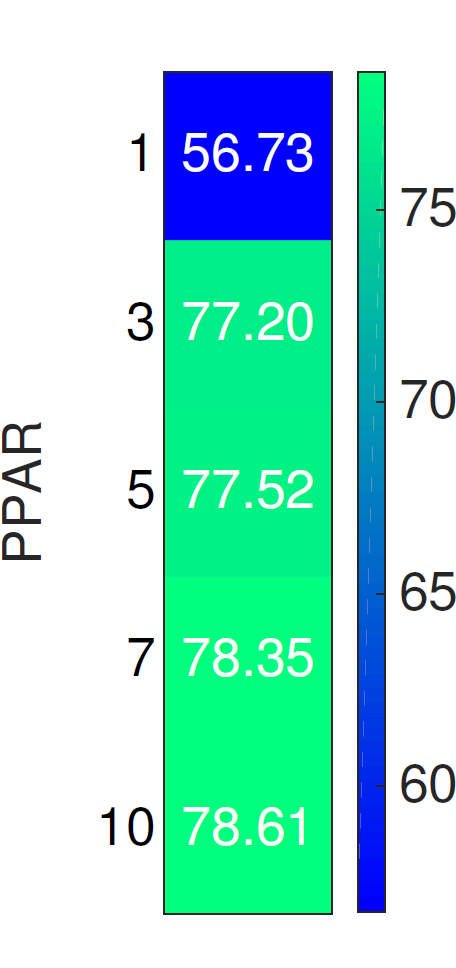}
		  \caption{ABSTRACT-50S}
          \label{fig:single_module_val}
    \end{subfigure}
	\begin{subfigure}[b]{0.36\textwidth}
		  \centering
		  \includegraphics[height=4.3cm]{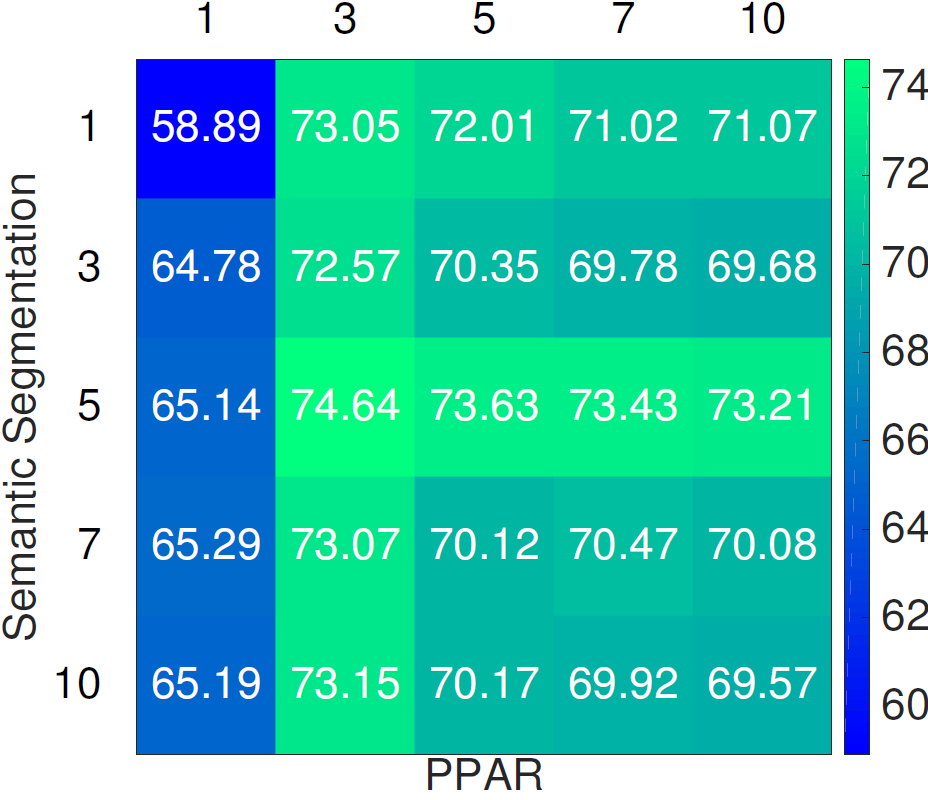}
		  \caption{PASCAL-50S}
          \label{fig:pascal_val}
    \end{subfigure}
    \begin{subfigure}[b]{0.36\textwidth}
    	  \centering
		  \includegraphics[height=4.3cm]{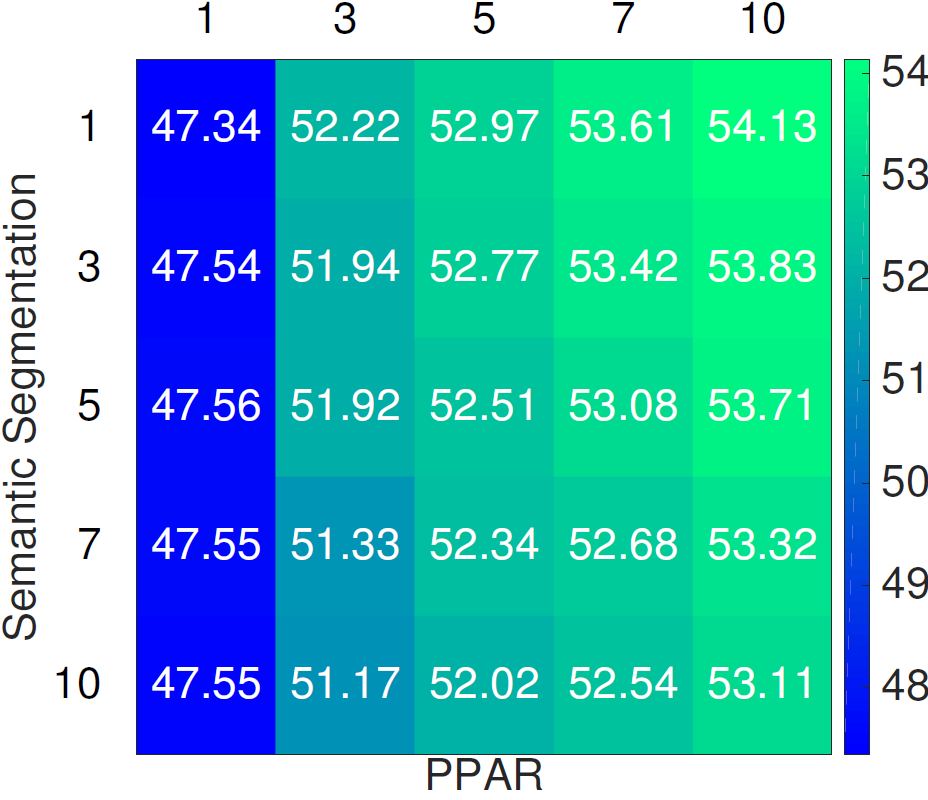}
		  \caption{PASCAL-Context-50S}
          \label{fig:pascal_context_val}
    \end{subfigure}
   \caption{{\small (a) Validation accuracies for different values of $M$ on ABSTRACT-50S, (b) for different values of $M_{\yb}, M_{\zb}$ on PASCAL-50S, (c) for different values of $M_{\yb}, M_{\zb}$ on PASCAL-Context-50S.}}
    \label{fig:captioned_val}
	\vspace{\captionReduceBot}
\end{figure*}

\begin{table*}[ht!]
\centering
\setlength{\tabcolsep}{12pt}
{\small
\scalebox{\tableScale}{
\begin{tabular}{@{\extracolsep{\fill}}p{3cm} ccc cccc@{\extracolsep{\fill}}}  
\toprule
 & \multicolumn{3}{c}{PASCAL-50S} & \multicolumn{3}{c}{PASCAL-Context-50S} \\ 
\cmidrule[0.75pt](lr){2-4}
\cmidrule[0.75pt](l){5-7}
 & \begin{tabular}[x]{@{}c@{}}Instance-Level \\ Jaccard Index \end{tabular} & PPAR Acc. & Average & \begin{tabular}[x]{@{}c@{}}Instance-Level \\ Jaccard Index \end{tabular} & PPAR Acc. & Average \\
\midrule
DeepLab-CRF
& 66.83 & - & -  & \textbf{43.94} & - & -\\
Stanford Parser 
& - & 62.42 & -  & - & 50.75 & - \\
Average & - & - & 64.63 & - & - & 47.345 \\
\midrule
Domain Adaptation & - & 72.08 & - & - & 58.32 & - \\
\midrule
Ours \cascade & 67.56 & 75.00 & 71.28 & \textbf{43.94} & \textbf{63.58} & \textbf{53.76} \\ 
Ours \mediator & \textbf{67.58} & \textbf{80.33} & \textbf{73.96} & \textbf{43.94} & \textbf{63.58} & \textbf{53.76} \\
\bottomrule
\oracle & 69.96 & 96.50 & 83.23 & 49.21 & 75.75 & 62.48 \\
\bottomrule
\end{tabular}
}
}
\caption{{\small Results on our subset of the PASCAL-50S and PASCAL-Context-50S datasets. We are able to significantly outperform the Stanford Parser
and make small improvements over DeepLab-CRF
for PASCAL-50S.}}
\label{table:sspp_exp_final}
\vspace{-0.4cm}
\end{table*}
 
We performed 10-fold cross-val for our results of PASCAL-50S and PASCAL-Context-50S, with 8 \train folds, 1 \val fold, and 1 \test fold, where the \val fold was used to pick $M_{\yb}$, $M_{\zb}$, and $C$.
\figref{fig:captioned_val} shows the average combined accuracy on \val, which was found to be maximal at $M_{\yb}= 5, M_{\zb} = 3$ for PASCAL-50S, and $M_{\yb}= 1, M_{\zb} = 10$ for PASCAL-Context-50S, which are used at test time.

We present our results in Table~\ref{table:sspp_exp_final}.
Our approach significantly outperforms the Stanford Parser~\cite{de2006generating} by  17.91\% (28.69\% relative) for PASCAL-50S, and 12.83\% (25.28\% relative) for PASCAL-Context-50S. We also make small  improvements over DeepLab-CRF~\cite{chen14semantic} in the case of PASCAL-50S.
To measure statistical significance of our results, we performed paired $t$-tests between \mediator and \mapmap. For both modules (and average), the null hypothesis (that the accuracies of our approach and baseline come from the same distribution) can be successfully rejected at p-value 0.05. For sake of completeness, we also compared \mediator with our ablated system (\cascade) and found statistically significant differences only in PPAR. 

\begin{figure*}[!ht]
\centering
\begin{minipage}[t]{.27\textwidth}
\centering
\vspace{0pt}
\includegraphics[width=1\columnwidth]{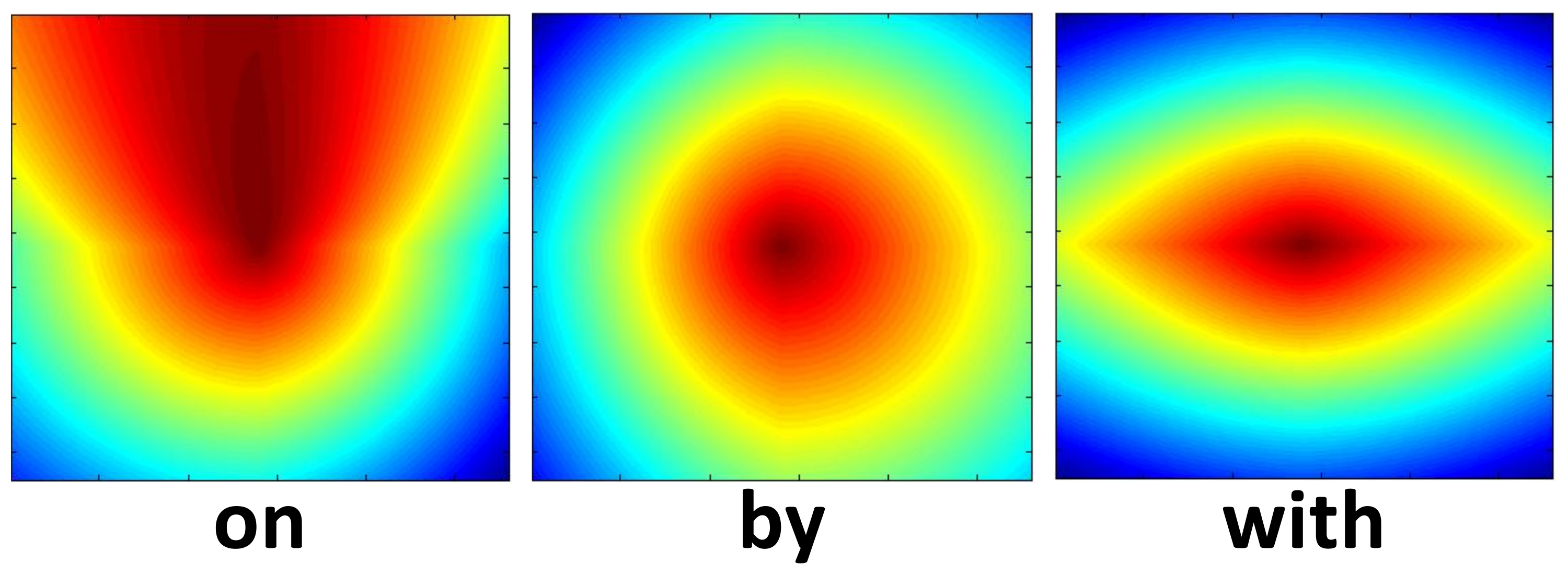}
\caption{{ Visualizations for 3 different prepositions (red = high scores, blue = low scores). We can see that our model has implicitly learned spatial arrangements unlike other spatial relation learning (SRL) works.}}
\label{fig:heat_maps}
\end{minipage}\hfill
\begin{minipage}[t]{.68\textwidth}
\centering
\vspace{0pt}
\setlength{\tabcolsep}{4pt}
{\small
\scalebox{\tableScale}{
\resizebox{\columnwidth}{!}{
\begin{tabular}{@{\extracolsep{\fill}}p{3.6cm} ccc c@{\extracolsep{\fill}}}   
\toprule
& \multicolumn{2}{c}{PASCAL-50S} & PASCAL-Context-50S \\
\cmidrule[0.75pt](lr){2-3}
\cmidrule[0.75pt](l){4-4}
 Feature set & \begin{tabular}[x]{@{}c@{}}Instance-Level \\ Jaccard Index \end{tabular} & PPAR Acc. & PPAR Acc. \\
\midrule
All features & 67.58 & 80.33 & 63.58\\ 
Drop all consistency & 66.96 & 66.67 & 61.47 \\ 
Drop Euclidean distance & 67.27 & 77.33 & 63.77 \\ 
Drop directional distance & 67.12 & 78.67 & 63.63 \\ 
Drop \wordtwovec  & 67.58 & 78.33 & 62.72 \\ 
Drop category presence & 67.48 & 79.25 & 61.19 \\  
\bottomrule
\end{tabular}
}
}
\captionof{table}{{\small Ablation study of different feature combinations. Only PPAR Acc. is shown for PASCAL-Context-50S because $M_y = 1$.}}
\label{table:ablation_features}
}
\end{minipage}
\vspace{-0.35cm}
\end{figure*}

These results demonstrate a need for each module to produce a diverse set of plausible hypotheses for our \mediator model to reason about. 
In the case of PASCAL-Context-50S, \mediator performs identical to \cascade since $M_{\yb}$ is chosen as 1 (which is the \cascade setting) in cross-validation. Recall that \mediator is a larger model class than \cascade (in fact, \cascade is a special case of \mediator with $M_y=1$). It is interesting to see that the large model class does not hurt, and \mediator gracefully reduces to a smaller capacity model (\cascade) if the amount of data is not enough to warrant the extra capacity. We hypothesize that in the presence of more training data, cross-validation may pick a different setting of $M_{\yb}$ and $M_{\zb}$, resulting in full utilization of the model capacity. 
Also note that our domain adaptation baseline achieved an accuracy higher than MAP/Stanford-Parser, 
but significantly lower than our approach for both PASCAL-50S and PASCAL-Context-50S.  We also performed this for our single-module experiment 
and picked $M_z$ (=10) with cross-validation, which resulted in an accuracy of 57.23\%. 
Again, this is higher than MAP/Stanford-Parser (56.73\%), but significantly lower than our approach (77.39\%). 
Clearly, domain adaptation alone is not sufficient.
We also see that \oracle performance is fairly high, suggesting that when there is ambiguity and room for improvement, \mediator is able to rerank effectively.

\textbf{Ablation Study for Features.} 
Table~\ref{table:ablation_features} displays results of an ablation study on PASCAL-50S and PASCAL-Context-50S to show the importance of the different features. In each row, we retain the module score features and drop a single set of consistency features. We can see all consistency features contribute to the performance of \mediator.

\textbf{Visualizing Prepositions.} 
 \figref{fig:heat_maps} shows a visualization for what our \mediator model has implicitly learned about 3 prepositions (``on'', ``by'', ``with''). These visualizations show the score obtained by taking the dot product of distance features (Euclidean and directional) between $object_1$ and $object_2$ connected by the preposition with the corresponding learned weights of the model, considering $object_2$ to be at the center of the visualization. Notice that these were learned without explicit training for spatial learning as in spatial relation learning (SRL) works~\cite{malinowski2014pooling, lan2012image}. These were simply recovered as an intermediate step towards reranking SS + PPAR hypotheses.
Also note that SRL cannot handle multiple segmentation hypotheses, which our work shows are important (Table~\ref{table:sspp_exp_final} \cascade). In addition, our approach is more general.

\vspace{\subsectionReduceTop}
\section{Discussions and Conclusion}
\vspace{\subsectionReduceBot}
\label{sec:conclusions}

We presented an approach to the simultaneous reasoning about prepositional phrase attachment resolution of captions and semantic segmentation in images
that integrates beliefs across the modules to pick the best pair of a diverse set of hypotheses. 
Our full model (\mediator) significantly improves the accuracy of PPAR over the Stanford Parser by 17.91\% for PASCAL-50S and by 12.83\% for PASCAL-Context-50S,
and achieves a small improvement on semantic segmentation over DeepLab-CRF for PASCAL-50S. 
These results demonstrate a need for information exchange between the modules, 
as well as a need for a diverse set of hypotheses to concisely capture the uncertainties of each module.  
Large gains in PPAR validate our intuition 
that vision is very helpful for dealing with ambiguity in language.
Furthermore, we see even larger gains are possible
from the oracle accuracies.

While we have demonstrated our approach on a task involving simultaneous reasoning about language and vision, our approach is general and can be used for other applications.  
Overall, we hope our approach will be useful in a number of settings.

\textbf{Acknowledgements.} 
We thank Larry Zitnick, Mohit Bansal, Kevin Gimpel, and Devi Parikh for helpful discussions, suggestions, and feedback included in this work. 
A majority of this work was done while AL was an intern at Virginia Tech. This work was partially supported by a National Science Foundation CAREER award, an Army Research Office YIP Award, an Office of Naval Research grant N00014-14-1-0679, and GPU donations by NVIDIA, all awarded to DB. 
GC was supported by a DTRA Basic Research grant HDTRA1-13-1-0015 provided by KK.
The views and conclusions contained herein are those of the authors and should not be interpreted as necessarily representing official policies or endorsements, either expressed or implied, of the U.S. Government or any sponsor. 

\clearpage

\appendix

\renewcommand{\thesection}{Appendix \Roman{section}} 

\section*{Appendix Overview} 

\vspace{-5pt}

In this appendix, we provide the following: 

\begin{compactitem}
\item[] \ref{sec:add_motivation}: Additional motivation for our \mediator model. 
\item[] \ref{sec:background_abstract}: Background on ABSTRACT-50S. 
\item[] \ref{sec:dataset_curation}: Details of the dataset curation process for the ABSTRACT-50S, PASCAL-50S, and PASCAL-Context-50S datasets.
\item[] \ref{sec:weighting_modules}: Results where we study the effect of varying the weighting of each module in our approach.
\item[] \ref{sec:prep_performance}: Performances and gains over the independent baseline on PASCAL-Context-50S for each preposition.
\item[] \ref{sec:qualitative_ex}: Qualitative examples from our approach.
\end{compactitem}

\section{Additional Motivation for \mediator}
\label{sec:add_motivation}

\begin{figure*}[ht!]
        \centering
    \begin{subfigure}[b]{0.3\linewidth}
		  \includegraphics[width=\linewidth]{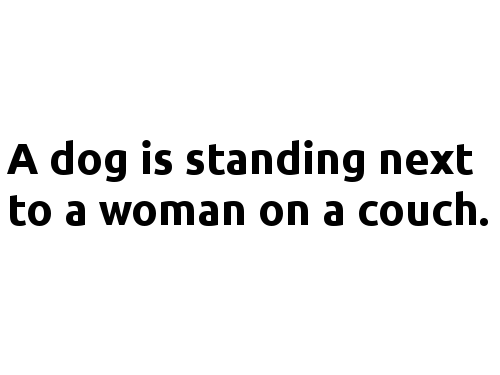}
		  \caption{Caption}
          \label{fig:pp_ex_sent}
    \end{subfigure}
	\begin{subfigure}[b]{0.3\linewidth}
		  \includegraphics[width=\linewidth]{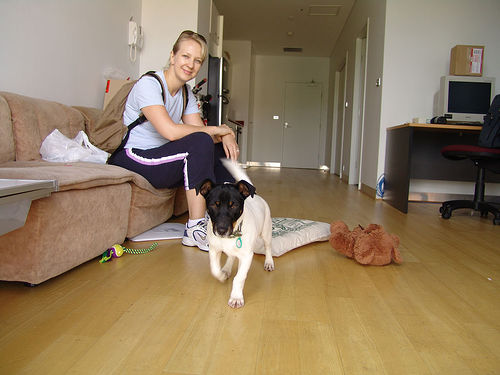}
		  \caption{Input Image}
          \label{fig:pp_ex_input}
    \end{subfigure}
    \begin{subfigure}[b]{0.3\linewidth}
		  \includegraphics[width=\linewidth]{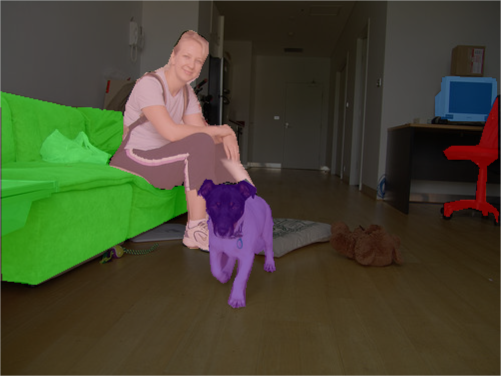}
		  \caption{Segmentation GT}
          \label{fig:pp_ex_gt}
    \end{subfigure}
	\begin{subfigure}[b]{0.3\linewidth}
		  \includegraphics[width=\linewidth]{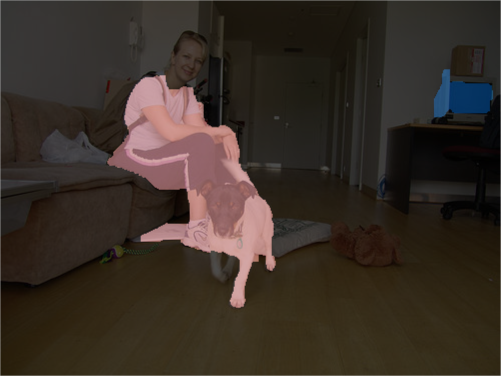}
		  \caption{Segmentation Hypothesis \#1}
          \label{fig:pp_ex_ss1}
    \end{subfigure}
 	\begin{subfigure}[b]{0.3\linewidth}
		  \includegraphics[width=\linewidth]{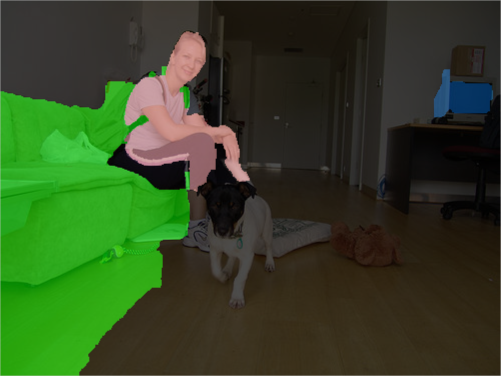}
		  \caption{Segmentation Hypothesis \#2}
          \label{fig:pp_ex_ss2}
    \end{subfigure}
 	\begin{subfigure}[b]{0.3\linewidth}
		  \includegraphics[width=\linewidth]{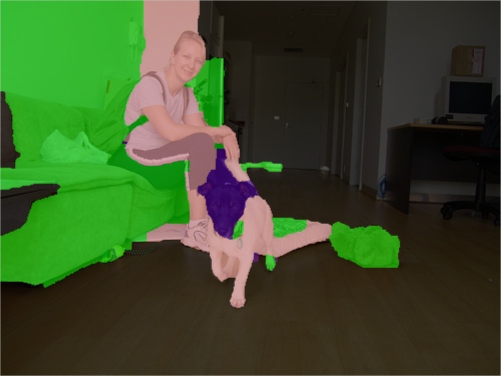}
		  \caption{Segmentation Hypothesis \#3}
          \label{fig:pp_ex_ss3}
    \end{subfigure}
	\begin{subfigure}[b]{0.3\linewidth}
		  \includegraphics[width=\linewidth]{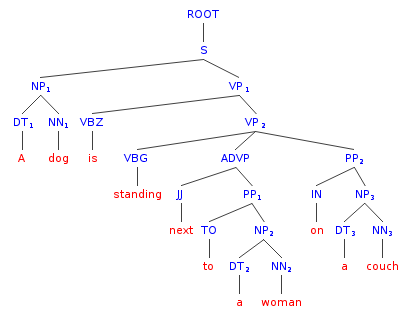}
		  \caption{Parse Hypothesis \#1}
          \label{fig:pp_ex_pp1}
    \end{subfigure}
 	\begin{subfigure}[b]{0.3\linewidth}
		  \includegraphics[width=\linewidth]{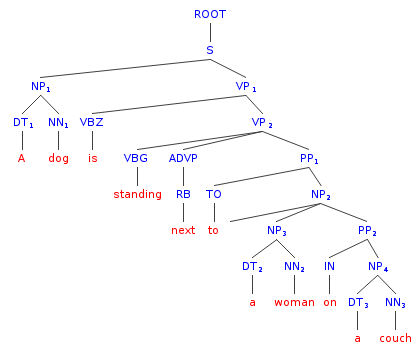}
		  \caption{Parse Hypothesis \#2}
          \label{fig:pp_ex_pp2}
    \end{subfigure}
 	\begin{subfigure}[b]{0.3\linewidth}
		  \includegraphics[width=\linewidth]{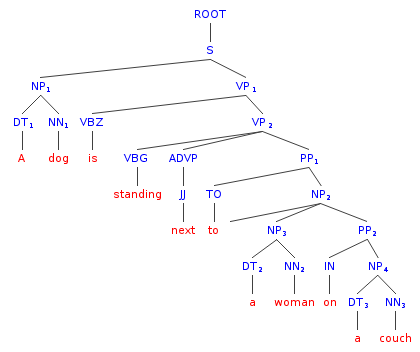}
		  \caption{Parse Hypothesis \#3}
          \label{fig:pp_ex_pp3}
    \end{subfigure}
    \caption{In this figure, we illustrate why the \mediator model makes sense for the task of captioned scene understanding. For the caption-image pair (\figref{fig:pp_ex_sent}-\figref{fig:pp_ex_input}), we see that parse tree \#1 (\figref{fig:pp_ex_pp1}) shows ``standing'' (the verb phrase of the noun ``dog'') connected with ``couch'' via the ``on'' preposition, whereas parse trees \#2 (\figref{fig:pp_ex_pp2}) and \#3 (\figref{fig:pp_ex_pp3}) show ``woman'' connected with ``couch'' via the ``on'' preposition. 
This ambiguity can be resolved if we look at an accurate semantic segmentation such as Hypothesis \#3 (\figref{fig:pp_ex_ss3}) of the associated image (\figref{fig:pp_ex_input}).
Likewise, we might be able to do better at semantic segmentation if we choose a segmentation 
that is consistent with the sentence, such as Segmentation Hypothesis \#3 (\figref{fig:pp_ex_ss3}), 
which contains a person on a couch with a dog next to them,
unlike the other two hypotheses (\figref{fig:pp_ex_ss1} and \figref{fig:pp_ex_ss2}).}
    \label{fig:pp_ex}
\end{figure*}

An example providing additional motivation for our approach is shown in \figref{fig:pp_ex}, where 
the ambiguous sentence that describes the image is ``A dog is standing next to a woman on a couch''. The ambiguity is ``(dog next to woman) on couch'' vs ``dog next to (woman on couch)'',
which is reflected in parse trees' uncertainty. 
Parse tree \#1 (\figref{fig:pp_ex_pp1}) shows ``standing'' (the verb phrase of the noun ``dog'') connected with ``couch'' via the ``on'' preposition, whereas parse trees \#2 (\figref{fig:pp_ex_pp2}) and \#3 (\figref{fig:pp_ex_pp3}) show ``woman'' connected with ``couch'' via the ``on'' preposition. 
This ambiguity can be resolved if we look at an accurate semantic segmentation such as Hypothesis \#3 (\figref{fig:pp_ex_ss3}) of the associated image (\figref{fig:pp_ex_input}).
Likewise, we might be able to do better at semantic segmentation if we choose a segmentation 
that is consistent with the sentence, such as Segmentation Hypothesis \#3 (\figref{fig:pp_ex_ss3}), 
which contains a person on a couch with a dog next to them,
unlike the other two hypotheses (\figref{fig:pp_ex_ss1} and \figref{fig:pp_ex_ss2}).

\section{Background About ABSTRACT-50S}
\label{sec:background_abstract}

\begin{figure}[h!]
\centering
\includegraphics[width=\columnwidth]{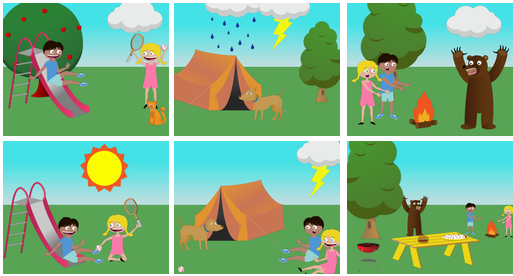}
\caption{We show some example scenes from \protect\cite{ZitnickCVPR2013}.
Each column shows two semantically similar scenes,
while the different columns show the diversity of scene types.}
\label{fig:clipart_examples}
\vspace{-1mm}
\end{figure}

The Abstract Scenes dataset~\cite{ZitnickCVPR2013} contains synthetic images 
generated by human subjects via a drag-and-drop clipart interface.
The subjects are given access to a (random) subset of 56 clipart objects that can be found
in park scenes, as well as two characters, Mike and Jenny, with a variety of poses and expressions.
Example scenes can be found in \figref{fig:clipart_examples}.
The motivation is to allow researchers to focus on higher-level semantic understanding
without having to deal with noisy information extraction from real images, since
the entire contents of the scene are known exactly, 
while also providing a dense semantic space to study 
(due to the heavily constrained world).
We used the dataset in precisely this way to first test out
the PPAR module in isolation to demonstrate that this
problem can be helped by a sentence's corresponding image.

\section{Dataset Curation and Annotation}
\label{sec:dataset_curation}

The subsets of the PASCAL-50S and ABSTRACT-50S datasets used in our experiments were carefully curated by two vision + NLP graduate students. The subset of the PASCAL-Context-50S dataset was curated by Amazon Mechanical Turk (AMT) workers. 
The following describes the dataset curation process for each datset.

\textbf{PASCAL-50S:} For PASCAL-50S we first obtained sentences that contain one or more of 7 prepositions
(\ie, ``with'', ``next to'', ``on top of'', ``in front of'', ``behind'', ``by'', and ``on'')
that intuitively would typically depend on the relative distance between objects. 
Then we look for sentences that have preposition phrase attachment ambiguities, 
\ie, sentences where the parser output has different sets of prepositions for different parsings. 
Due to our focus on PP attachment,
we do not pay attention to other parts of the sentence parse, 
so the parses can change while the PP attachments remain the same,
as in \figref{fig:pp_ex_pp2} and \figref{fig:pp_ex_pp3}. 
The sentences thus obtained are further filtered to obtain sentences in which the objects 
that are connected by the preposition belonging to one of the 20 PASCAL object categories. 
Since our vision module is semantic segmentation and not instance-level segmentation, we restrict the dataset to sentences involving prepositions connecting two different PASCAL categories. 
Thus, our final dataset contains 100 sentences describing 30 unique images and 
contains 16 of the 20 PASCAL categories as described in the paper. We then manually annotated the ground truth PP attachments.
Such manual labeling by student annotators with expertise in NLP takes a lot of time, but results in annotations that are linguistically high-quality, with any inter-human disagreement resolved by strict adherence to rules of grammar.

\textbf{ABSTRACT-50S:} 
We first obtained sentences that contain one or more of 6 prepositions 
(\ie, ``with'', ``next to'', ``on top of'', ``in front of'', ``behind'', ``under'').
Due to the semantic differences between the datasets, not all prepositions found
in one were present in the other. 
Further filtering on sentences was done to ensure that the sentences contain at least one 
preposition phrase attachment ambiguity that is between the clipart noun categories 
(\ie, each clipart piece has a name, like ``snake'', that we search the sentence parsing for).
This filtering reduced the original dataset of 25,000 sentences and 500 scenes 
to our final experiment dataset of 399 sentences and 201 scenes.
We then manually annotated the ground truth PP attachments.

\begin{figure}[h!]
\centering
\includegraphics[width=\columnwidth]{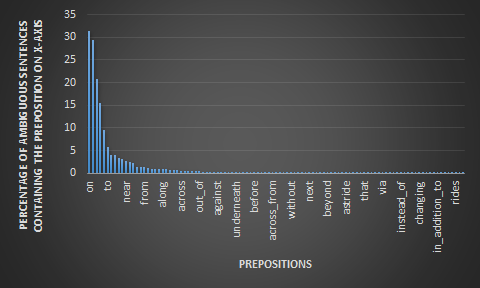}
\caption{We show the percentage of ambiguous sentences in PASCAL-Context-50S dataset before filtering for prepositions. We found that there was a drop in the percentage of sentences for prepositions that appear in the sorted list after ``down". So, for the PASCAL-Context-50S dataset we only keep sentences that have one or more visual prepositions in the list of prepositions up to ``down".}
\label{fig:prep_plot}
\end{figure}

\textbf{PASCAL-Context-50S}: For PASCAL-Context-50S, we first selected all sentences that have preposition phrase attachment ambiguities. We then plotted the distribution of prepositions in these sentences (see \figref{fig:prep_plot}). We found that there was a drop in the percentage of sentences for prepositions that appear in the sorted list after ``down". Therefore, we only kept sentences that have one or more 2-D visual prepositions in the list of prepositions up to ``down". Thus we ended up with the following 7 prepositions: ``on'', ``with'', ``next to'', ``in front of'', ``by'', ``near'', and ``down''. We then further sampled sentences to ensure uniform distribution across prepositions. Unlike PASCAL-50S, we did not filter sentences based on whether the objects connected by the prepositions belong to one of 60 PASCAL Context categories or not. Instead, we used the \wordtwovec~\cite{mikolov2013efficient} similarity between the objects in the sentence and the PASCAL Context categories as one of the features. Thus, our final dataset contains 1,822 sentences describing 966 unique images.

\begin{figure*}[ht!]
\centering
\includegraphics[width=\linewidth]{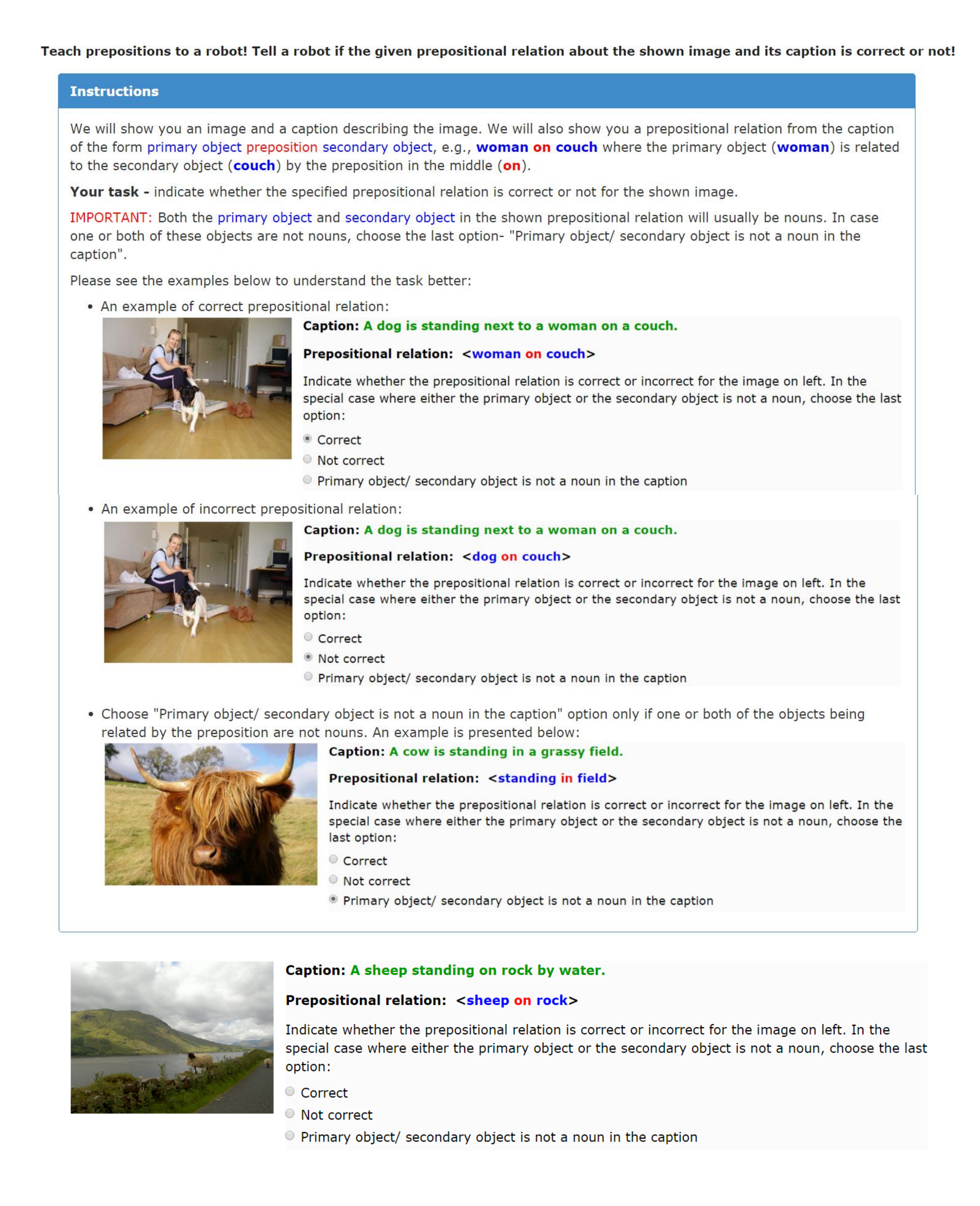}
\caption{The AMT interface to collect ground truth annotations for prepositional relations. Five answers were collected for each prepositional relation. The majority response is used for evaluation. The AMT workers are asked to select if the preposition is correct, not correct, or that the primary or secondary object is not a noun in the caption. Examples for all three answer choices are shown in the instructions presented to the workers.}
\label{fig:mturk}
\end{figure*}

The ground truth PP attachments for these 1,822 sentences were annotated by AMT workers. For each unique prepositional relation in a sentence, we showed the workers the prepositional relation of the form \textcolor{blue}{primary object} \textcolor{red}{preposition} \textcolor{blue}{secondary object} and its associated image and sentence and asked them to specify whether the prepositional relation is correct or not correct. We also asked them to choose the third option - ``Primary object/ secondary object is not a noun in the caption'' in case that happened. The user interface used to collect these annotations is shown in \figref{fig:mturk}. We collected five answers for each prepositional relation. For evaluation, we used the majority response. We found that 87.11\% of human responses agree with the majority response, indicating that even though AMT workers were not explicitly trained in rules of grammar by us, there is relatively high inter-human agreement.

\begin{table*}[ht!]
\centering
\setlength{\tabcolsep}{17pt}
{\small
\scalebox{\tableScale}{
\begin{tabular}{cccccccc}
\toprule  
& ``on'' & ``with'' & ``next to'' & ``in front of'' & ``by'' & ``near'' & ``down'' \\
\midrule
Acc. & 64.26 & 63.30 & 60.98 & 56.86 & 62.81 & 67.83 & 67.23 \\
Gain & 12.47 & 15.41 & 11.18 & 13.27 & 14.13 & 12.82 & 17.16 \\
\bottomrule
\end{tabular}
}
}
\caption{{\small Performances and gains over the independent baseline on PASCAL-Context-50S for each preposition.}}
\label{table:prep_acc}
\end{table*}

\section{Effect of Different Weighting of Modules} 
\label{sec:weighting_modules}

\begin{figure}[h!]
	\begin{center}
	\begin{subfigure}[b]{0.48\textwidth}
			\includegraphics[width=\columnwidth]{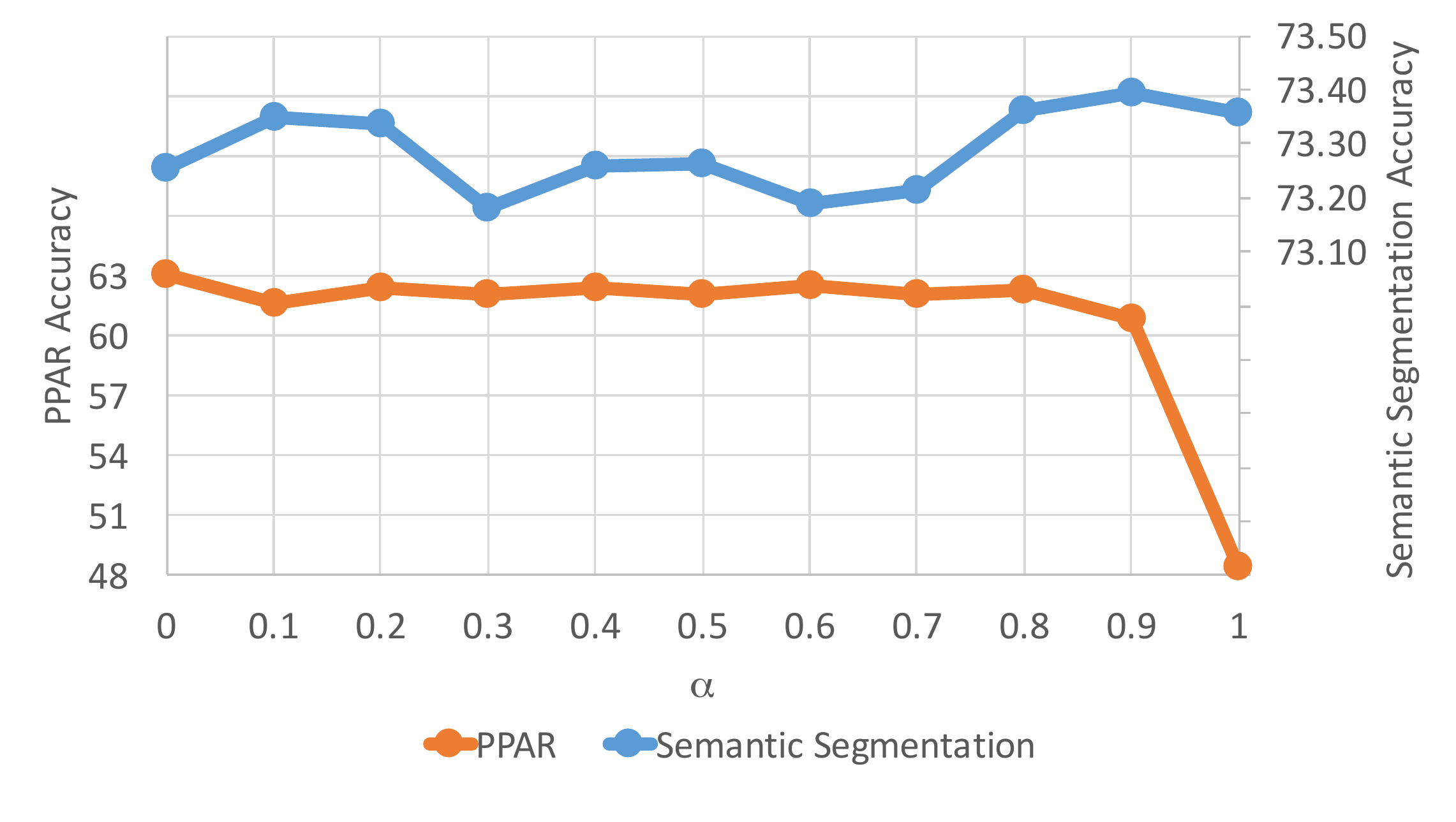}
			\caption{PASCAL-50S}
			\label{fig:pp_alpha_50S}
    \end{subfigure}
    \begin{subfigure}[b]{0.48\textwidth}
			\includegraphics[width=\columnwidth]{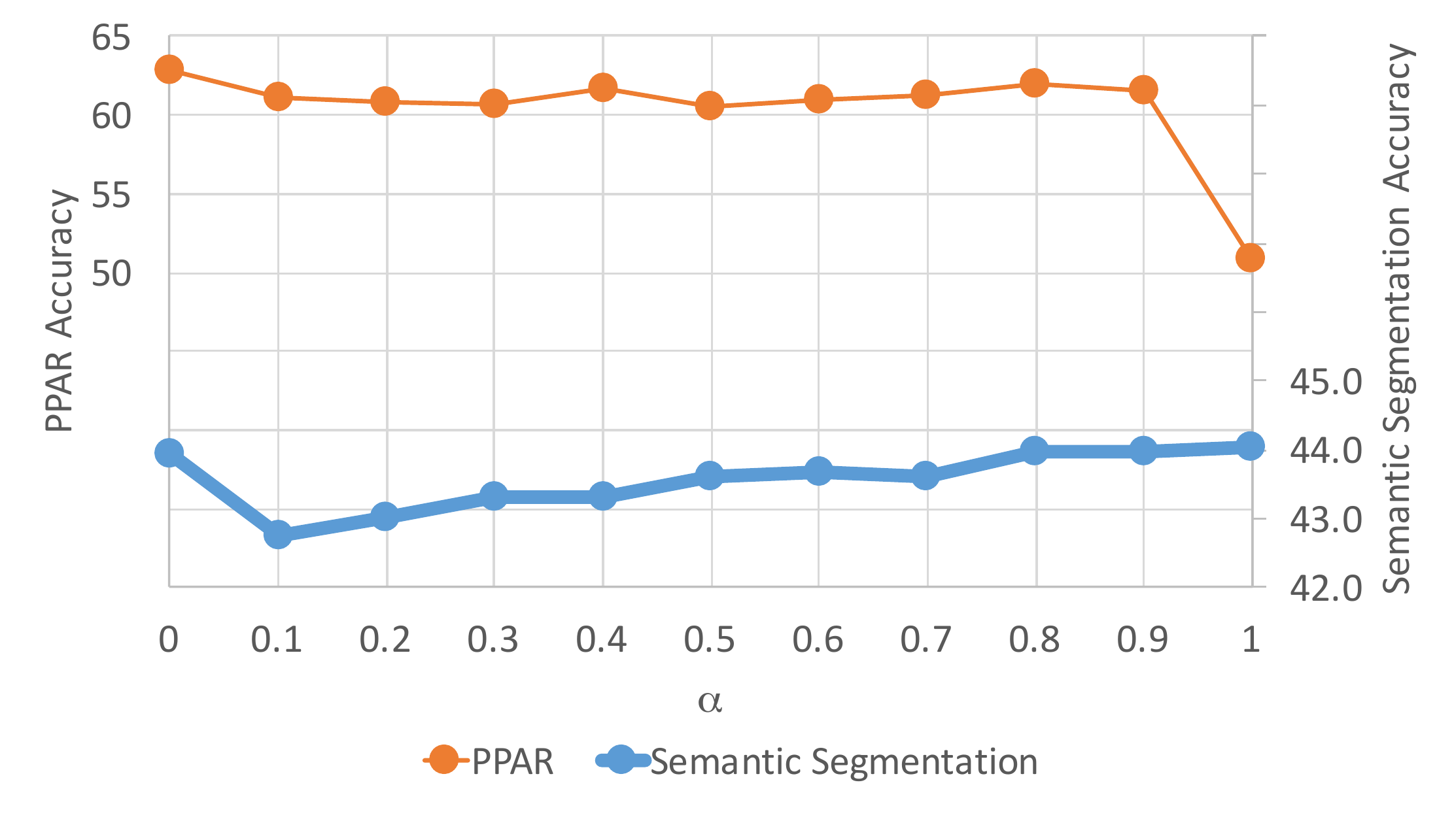}
			\caption{PASCAL-Context-50S}
			\label{fig:pp_alpha_pascal_context}
    \end{subfigure}
	\captionof{figure}{Accuracies \mediator (both modules) vs $\alpha$, where $\alpha$ is the coefficient for the 
	semantic segmentation module and 1-$\alpha$ is the coefficient for the PPAR resolution module in the loss function. Our approach is fairly robust to the setting of $\alpha$, as long as it is not set to either extremes, since that limits the synergy between the modules. (a) shows the results for PASCAL-50S, and (b) shows the results for PASCAL-Context-50S.}
	\label{fig:pp_alpha}
	\end{center}
\end{figure}

So far we have used the ``natural'' setting of $\alpha=0.5$, which gives equal weight to both modules. Note that $\alpha$ is not a parameter of our approach; it is a design choice that the user/experiment-designer makes. To see the effect of weighting the modules differently, we tested our approach for various values of $\alpha$.
\figref{fig:pp_alpha} shows how the accuracies of each module vary depending on $\alpha$ for the \mediator model for PASCAL-50S and PASCAL-Context-50S.
Recall that
$\alpha$ is the coefficient for the semantic segmentation module 
and 1-$\alpha$ is the coefficient for the PPAR resolution module in the loss function.
We see that as expected, putting no or little weight on the PPAR module drastically hurts performance for that module.
Our approach is fairly robust to the setting of $\alpha$, with a peak lying 
but any weight on it performs fairly similar with the peak lying somewhere between the extremes.
The segmentation module
has similar behavior, though it is not as sensitive to the choice of $\alpha$. 
We believe this is because of small ``dynamic range'' of this module -- the gap between the 1-best and oracle segmentation is smaller and thus the \mediator can always default to the 1-best as a safe choice.

\section{Performances for Each Preposition} 
\label{sec:prep_performance}

We provide performances and gains over the independent baseline on PASCAL-Context-50S for each preposition in Table~\ref{table:prep_acc}. We see that vision helps all prepositions.

\section{Qualitative Examples}
\label{sec:qualitative_ex}

\begin{figure*}[ht!]
	\centering
	\includegraphics[width=0.7\textwidth]{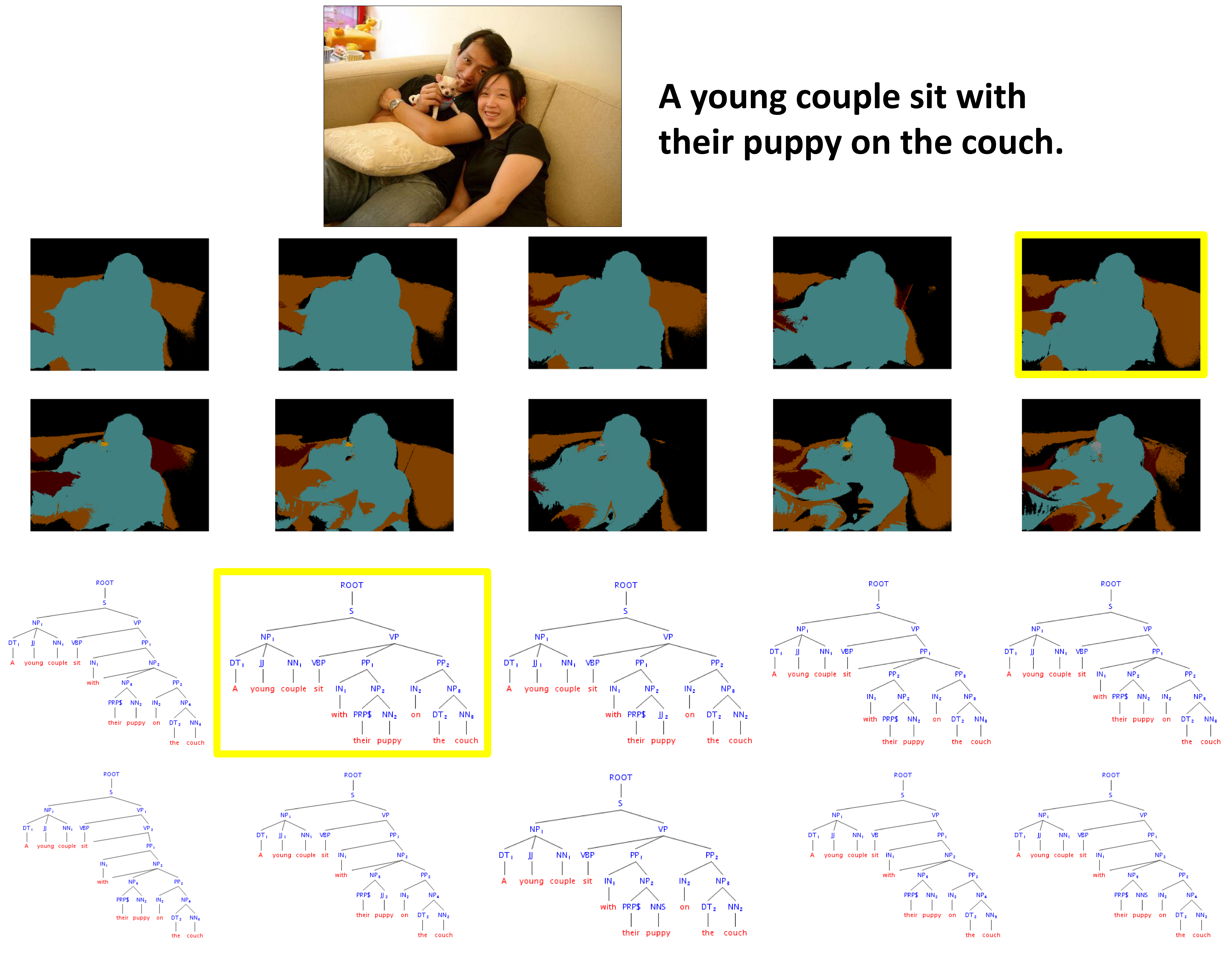}
	\caption{Example 1 -- multiple modules (SS and PPAR).}
	\label{fig:pp_qual_1}
\end{figure*}

\begin{figure*}[ht!]
	\centering
	\includegraphics[width=0.7\textwidth]{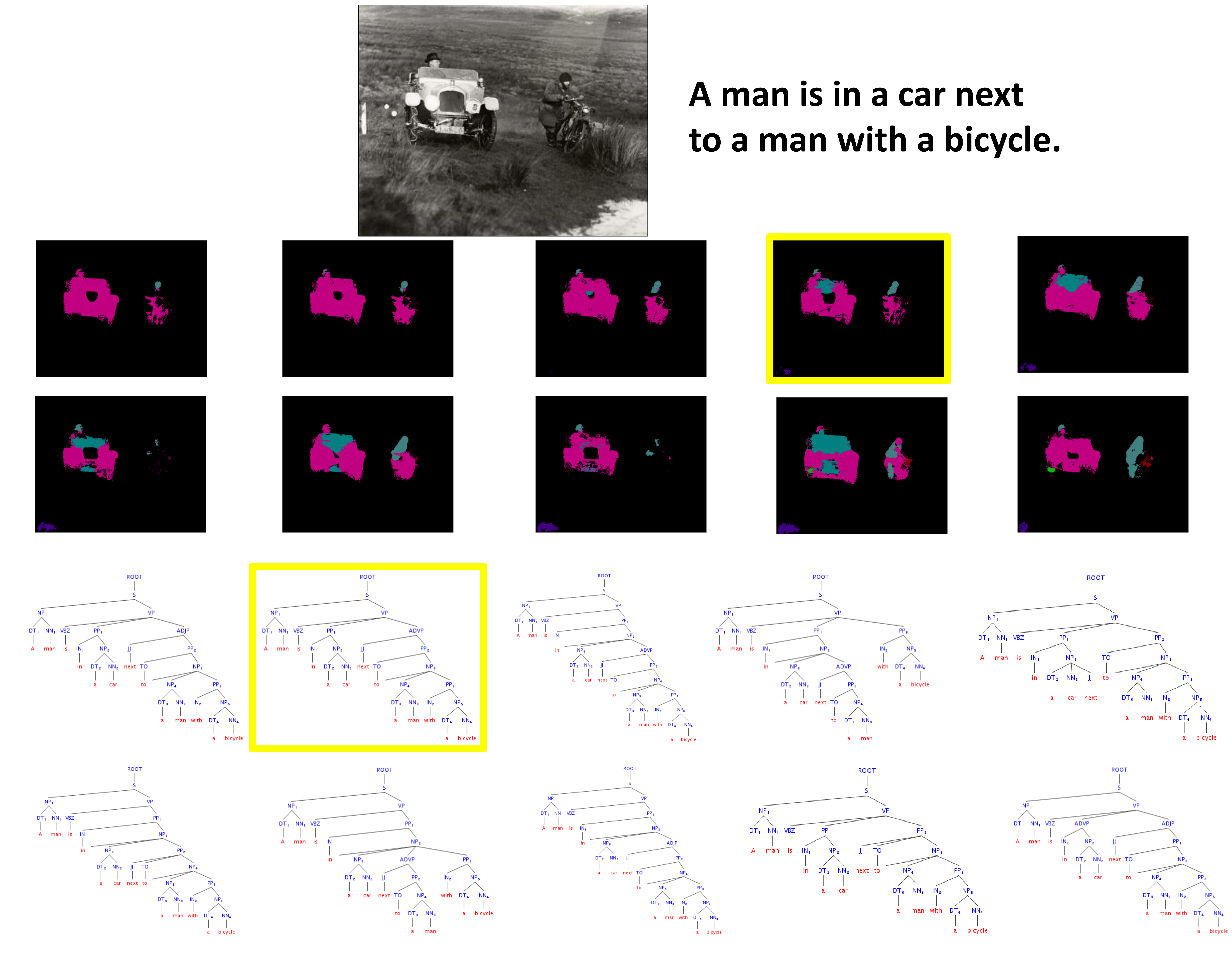}
	\caption{Example 2 -- multiple modules (SS and PPAR).}
	\label{fig:pp_qual_2}
\end{figure*}

\begin{figure*}[ht!]
	\centering
	\includegraphics[width=0.7\textwidth]{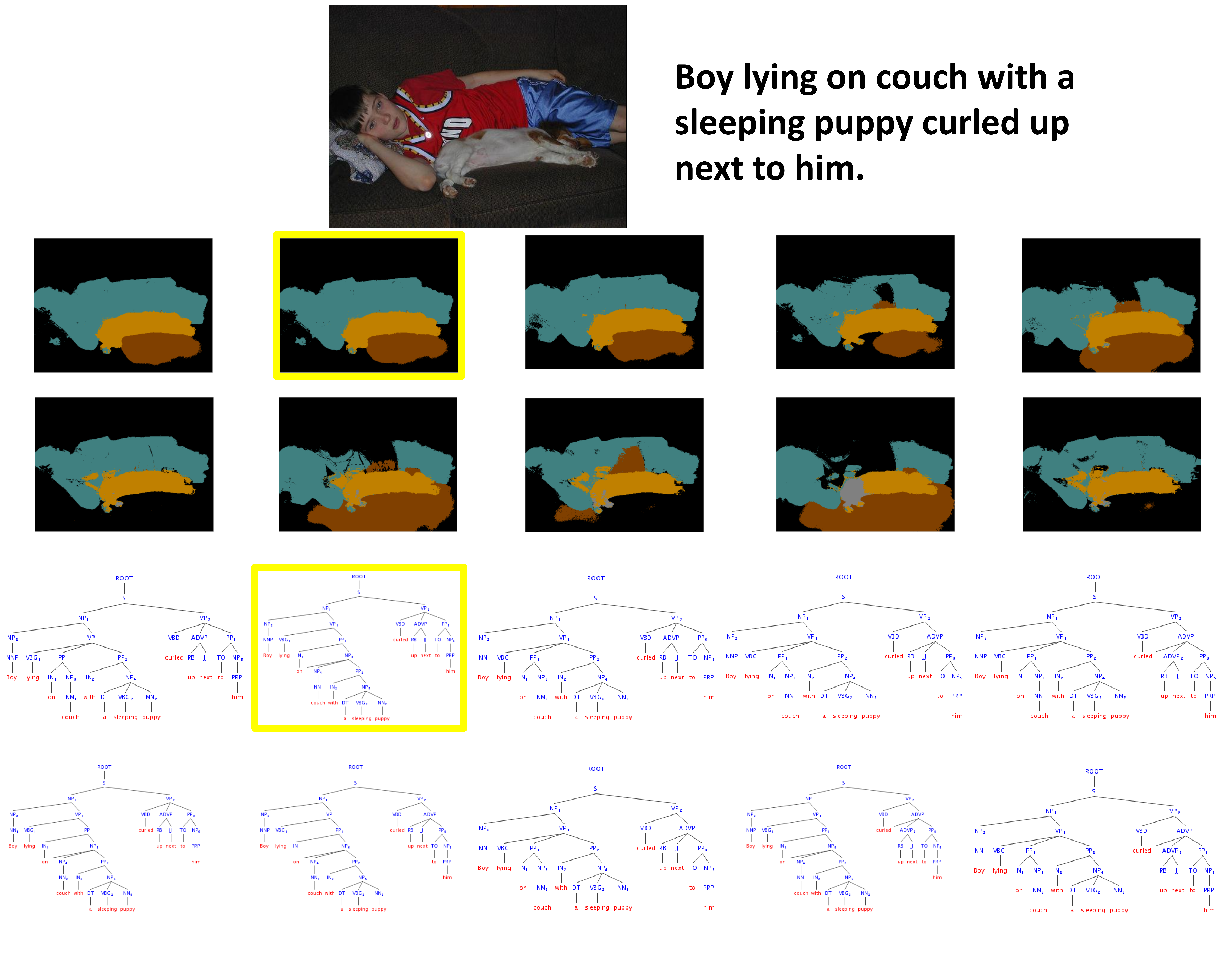}
	\caption{Example 3 -- multiple modules (SS and PPAR).}
	\label{fig:pp_qual_3}
\end{figure*}

\begin{figure*}[ht!]
	\centering
	\includegraphics[width=0.8\textwidth]{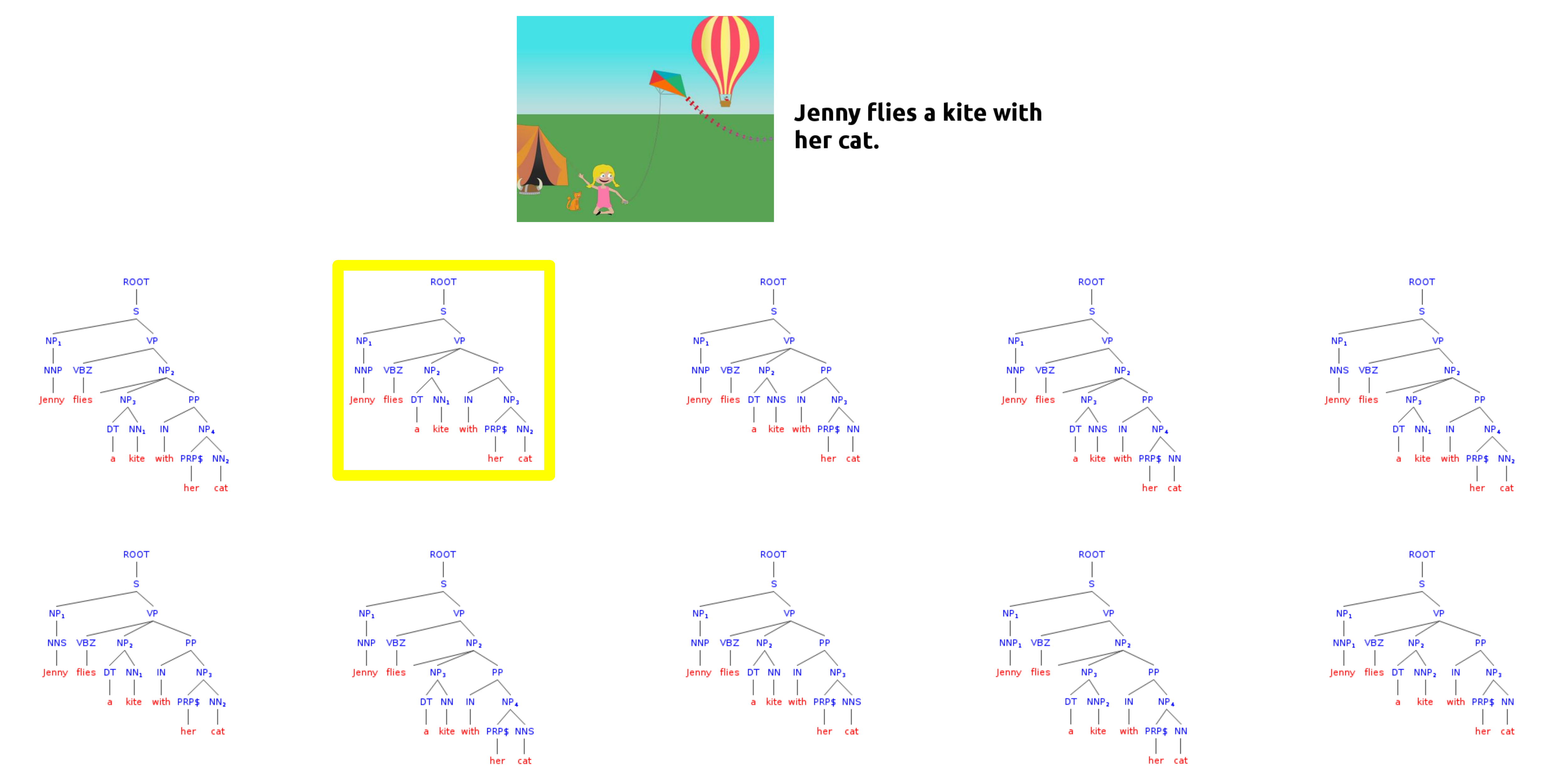}
	\caption{Example 1 -- single module (PPAR).}
	\label{fig:pp_qual_clipart_1}
\end{figure*}

\begin{figure*}[ht!]
	\centering
	\includegraphics[width=0.8\textwidth]{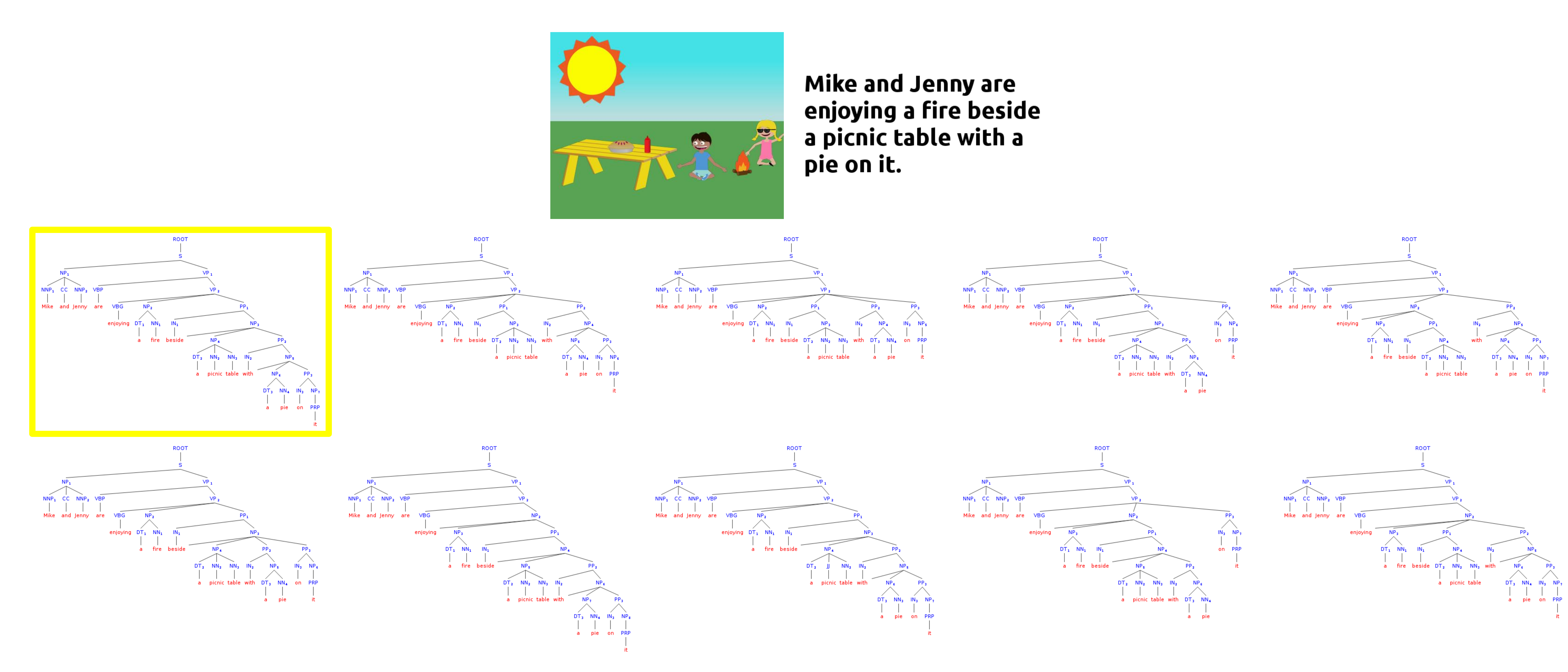}
	\caption{Example 2 -- single module (PPAR).}
	\label{fig:pp_qual_clipart_2}
\end{figure*}

\begin{figure*}[ht!]
	\centering
	\includegraphics[width=0.8\textwidth]{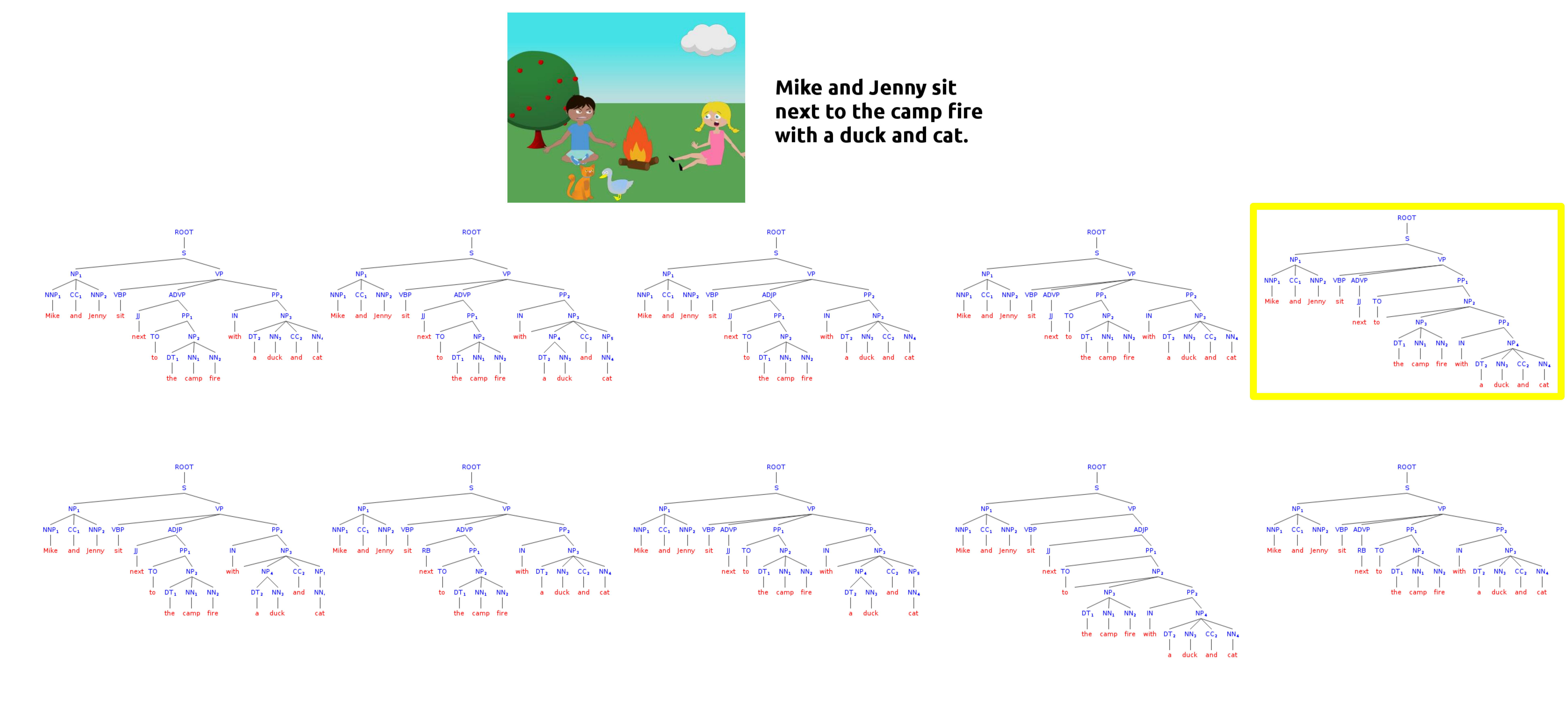}
	\caption{Example 3 -- single module (PPAR).}
	\label{fig:pp_qual_clipart_3}
\end{figure*}

\figref{fig:pp_qual_1} - \figref{fig:pp_qual_clipart_3} show qualitative examples for our experiments.  \figref{fig:pp_qual_1} - \figref{fig:pp_qual_3} show examples for the multiple modules examples (semantic segmentation and PPAR), and \figref{fig:pp_qual_clipart_1} - \figref{fig:pp_qual_clipart_3} show examples for the single module experiment. In each figure, the top row shows the image and the associated sentence. For the multiple modules figures, the second and third row show the diverse segmentations of the image, and the bottom two rows show different parsings of the sentence (last two rows for single module examples, as well).  In these examples our approach uses 10 diverse solutions for the semantic segmentation module and 10 different solutions for the PPAR module. The highlighted pairs of solutions show the solutions picked by the \mediator model.  Examining the results can give you a sense of how the parsings can help the semantic segmentation module pick the best solution and vice-versa.

\clearpage

\clearpage

{\small
\bibliographystyle{emnlp2016}
\bibliography{dbatra,gordon_bib_file,laddha_bib_file,aishwarya_bib_file}
}

\end{document}